\documentclass[10pt,twocolumn,letterpaper]{article}

\usepackage{wacv}
\usepackage{times}
\usepackage{epsfig}
\usepackage{graphicx}
\usepackage{amsmath}
\usepackage{amssymb}
\usepackage{booktabs}

\usepackage[accsupp]{axessibility}

\usepackage{booktabs}
\usepackage{array}
\newcolumntype{L}[1]{>{\raggedright\let\newline\\\arraybackslash\hspace{0pt}}m{#1}}
\newcolumntype{C}[1]{>{\centering\let\newline\\\arraybackslash\hspace{0pt}}m{#1}}
\newcolumntype{R}[1]{>{\raggedleft\let\newline\\\arraybackslash\hspace{0pt}}m{#1}}
\usepackage{tablefootnote}
\usepackage[font=small]{caption}
\usepackage{subcaption}
\usepackage{multirow}
\usepackage{enumitem}
\usepackage{float}

\usepackage{makecell}
\usepackage[toc,page]{appendix}

\makeatletter
\newcommand\xleftrightarrow[2][]{%
  \ext@arrow 9999{\longleftrightarrowfill@}{#1}{#2}}
\newcommand\longleftrightarrowfill@{%
  \arrowfill@\leftarrow\relbar\rightarrow}
\newcommand\xLeftrightarrow[2][]{%
  \ext@arrow 9999{\Longleftrightarrowfill@}{#1}{#2}}
\newcommand\Longleftrightarrowfill@{%
  \arrowfill@\Leftarrow\relbar\Rightarrow}
\makeatother

\usepackage{xcolor}
\newcommand{\best}[1]{
    \bf{#1}
}

\usepackage[pagebackref=true,breaklinks=true,colorlinks,bookmarks=false]{hyperref}

\def\wacvPaperID{504} 

\wacvalgorithmstrack   

\wacvfinalcopy 

\ifwacvfinal
\usepackage[breaklinks=true,bookmarks=false]{hyperref}
\else
\usepackage[pagebackref=true,breaklinks=true,colorlinks,bookmarks=false]{hyperref}
\fi

\begin{document}

\title{Image Completion with Heterogeneously Filtered Spectral Hints}

\author{
    Xingqian Xu$^1$, Shant Navasardyan$^3$, Vahram Tadevosyan$^3$, 
    Andranik Sargsyan$^3$, 
    \\Yadong Mu$^2$, and Humphrey Shi$^{1,3}$
    \\
    \\{\small$^1$SHI Lab @ UIUC \& UO, $^2$Peking University, $^3$Picsart AI Research (PAIR)}
}


\twocolumn[{
\maketitle
\begin{center}
    \vspace{-0.8cm}
    \captionsetup{type=figure}
    \includegraphics[width=0.8\textwidth]{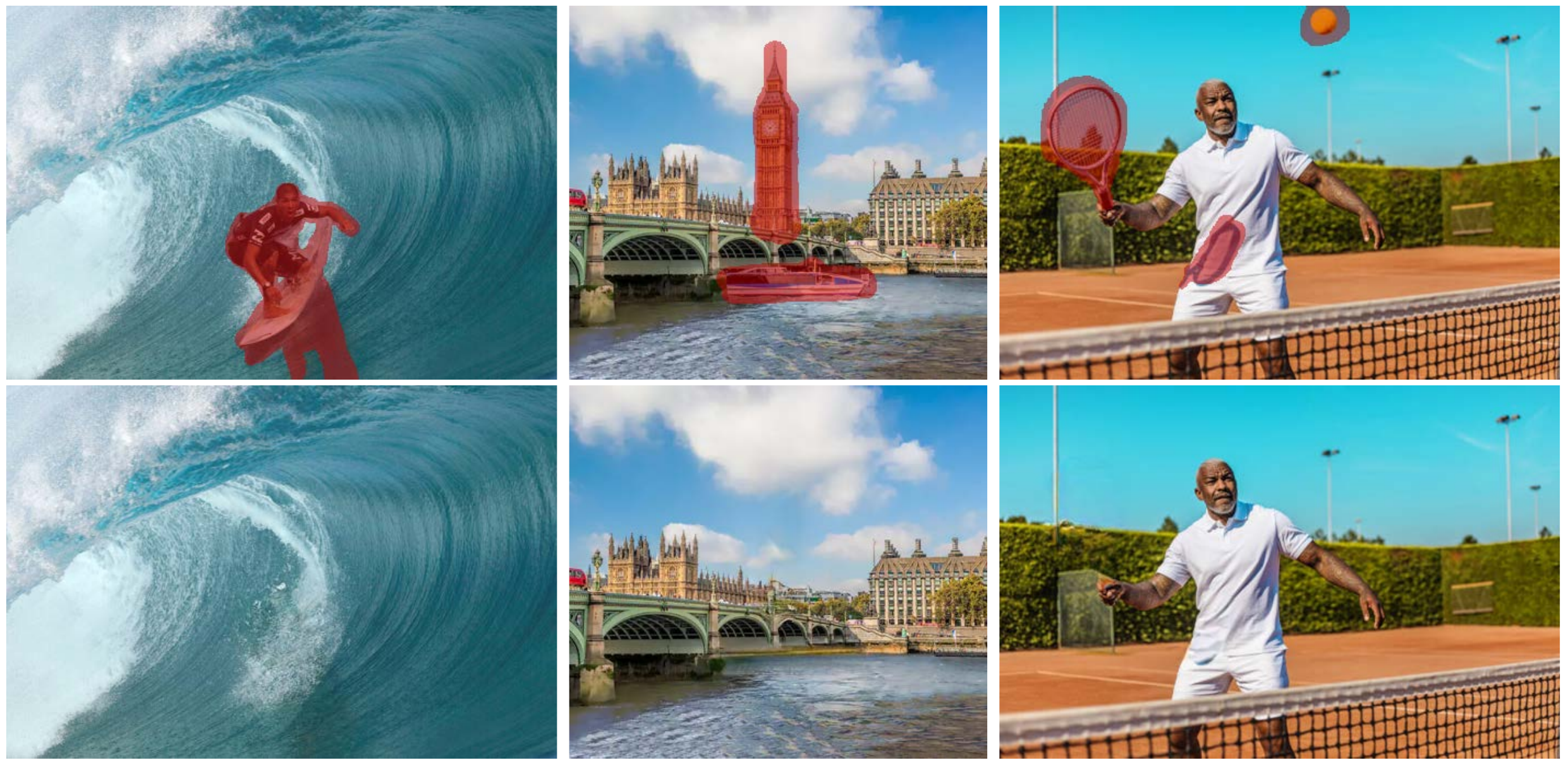}
    \captionof{figure}{Results of SH-GAN on a variety of inpainting cases. SH-GAN can fill-in image content with outstanding consistency.}
    \label{fig:teasor}
    \vspace{0.2cm}
\end{center}
}]
\thispagestyle{empty}


\begin{abstract}

\vspace{-0.5cm}

Image completion with large-scale free-form missing regions is one of the most challenging tasks for the computer vision community. While researchers pursue better solutions, drawbacks such as pattern unawareness, blurry textures, and structure distortion remain noticeable, and thus leave space for improvement. To overcome these challenges, we propose a new StyleGAN-based image completion network, \textit{Spectral Hint GAN} (\textbf{SH-GAN}), inside which a carefully designed spectral processing module, \textit{Spectral Hint Unit}, is introduced. We also propose two novel 2D spectral processing strategies, \textit{Heterogeneous Filtering} and \textit{Gaussian Split} that well-fit modern deep learning models and may further be extended to other tasks. From our inclusive experiments, we demonstrate that our model can reach FID scores of 3.4134 and 7.0277 on the benchmark datasets FFHQ and Places2, and therefore outperforms prior works and reaches a new state-of-the-art. We also prove the effectiveness of our design via ablation studies, from which one may notice that the aforementioned challenges, \ie pattern unawareness, blurry textures, and structure distortion, can be noticeably resolved.
Our code will be open-sourced at:
\href{https://github.com/SHI-Labs/SH-GAN}{https://github.com/SHI-Labs/SH-GAN}.


\end{abstract}


\vspace{-0.5cm}

\section{Introduction}

\vspace{-0.3cm}

Spectral analysis is a well-established research topic and has been intensely studied for decades~\cite{ft-book2,ft-book1,ft-book0,ft-article0}. Its corresponding downstream techniques in remote sensing, telecommunication, and healthcare significantly affect our modern life. Earlier computer vision techniques largely adopted algorithms such as the Fourier transform~\cite{ft-huang0,ft0}, wavelet transform~\cite{wt-denoise1,wt-denoise0,wt-raster0,wt-huang0} and curvelet transform~\cite{ct-denoise0} for image denoising, anti-aliasing, restoration, and compression. Ever since the rapid growth of deep learning, spectral analysis on images has fallen from popularity largely due to the fact that the intriguing properties of the 2D frequency domain complicated many solutions for the content-based tasks. Nevertheless, as the progress of deep learning research goes deeper and wider, researchers start to re-focus on the image spectral analysis and its potential applications. Recent works, such as ~\cite{ssd-gan,ffc,specbias,lama,ultrasr,syncspeccnn,wt-styletrans0,wt-sr}, have shown that a network structure with spectral priors can be favorable in many tasks, including classification, segmentation, image synthesis, and super-resolution. These works will undoubtedly guide the future computer vision research in spectral analysis. 

Despite the fast development of spectral-related deep learning methods, few works~\cite{ssd-gan,stylegan3,lama} have explored the potential of image completion with spectral priors, in which the major purpose of~\cite{ssd-gan,stylegan3} is still image synthesis. In the past few years, image completion heavily relied on feature extraction using CNNs and similarity-based patch matching techniques ~\cite{onionconv,hifill,deepfillv1,deepfillv2,crfill}. While these strategies have been proved useful in some scenarios, it remains a case-dependent approach due to its frequently shown structural distortion and texture artifacts. Meanwhile, the success of the StyleGAN series~\cite{stylegan-ada,stylegan3,stylegan1,stylegan2} on generation tasks has established a robust baseline for many downstream tasks such as style transfer~\cite{styleflow,stylegan-nada,styleclip}, GAN-inverse~\cite{dgp}, latent space editing~\cite{ganspace,faceedit,stylerig}, and inpainting~\cite{comodgan}. Among them, the recent image completion work CoModGAN~\cite{comodgan} introduced the concept of co-modulation and pushed the performance to the next level. 

Through our research, we compared several image completion works such as DeepFill~\cite{deepfillv1,deepfillv2}, LaMa~\cite{lama}, CoModGAN~\cite{comodgan}, \etc. We noticed that these methods produce promising results in some cases but struggle in others.
For instance, the patch-based approaches keep a good texture consistency for irregular textures with small textons (\eg grass, wood, asphalt, \etc), but create large structural distortions, especially when the unknown masks are large. LaMa, on the other hand, utilizes a spectral transform module, namely FFC~\cite{ffc}, and stands out in cases with strong pattern-like signals. However, it faces challenges in complex scenes and tends to create fast-repeating artifacts that are conceptually blurry. CoModGAN generates more natural-looking image content. However, CoModGAN is less bounded by image known regions; thus, it may ignore global patterns and create faulty objects. 

Motivated by the prior works using spectral transform to handle low-level patterns and modulated generative block to handle high-level semantic, we introduce a novel approach: \textit{Spectral Hint GAN} (SH-GAN) along with a new module: \textit{Spectral Hint Unit} (SHU). Our goal is to minimize the summarized problems, \ie, pattern unawareness, blurry textures, and structural distortions and to maintain the natural balance between pattern and semantic consistency. Moreover, we also propose two new spectral transform strategies: \textit{Heterogeneous Filtering}, and \textit{Gaussian Split}. Heterogeneous Filtering aims to manipulate the 2D spectrum using a learnable smooth-varying function correlated with its frequency. Meanwhile, Gaussian Split is a spectral split algorithm that distributes information to different resolution scales for image synthesis. As a result, our FID performances are 3.4134 on FFHQ~\cite{stylegan1} and 7.0277 on Places2~\cite{places2}, outperforming CoModGAN and other prior works and reaching a new state-of-the-art. 
We also demonstrate that our model outperforms LaMa and the other works on the narrow and wide masks using a LaMa-style evaluation scheme. Lastly, we perform ablation studies on SH-GAN, from which we clearly see the performance gain using SHU and the proposed spectral transform strategies, \ie Heterogeneous Filtering and Gaussian Split.

\vspace{0.1cm}
In summary, the main contributions of our work are the following:
\vspace{-0.3cm}

\begin{itemize}
    \setlength\itemsep{-0.2em}
    \item We propose a novel spectral-aware StyleGAN-based image completion network, \textit{Spectral Hint GAN} (SH-GAN), in which a new module, \textit{Spectral Hint Unit} (SHU), is introduced.
    \item We also bring out two new spectral processing strategies: \textit{Heterogeneous Filtering} and \textit{Gaussian Split}. These strategies aim to enhance the 2D spectral signals and hierarchically fuse them inside the synthesis network.
    \item The FID scores of SH-GAN outperform the state-of-the-art on two popular benchmark datasets: FFHQ and Places2. Meanwhile, we perform inclusive studies, through which we demonstrate the effectiveness of our new designs.
\end{itemize}

\section{Related Works}
\begin{figure*}[t]
    \centering
    \includegraphics[width=0.8\textwidth]{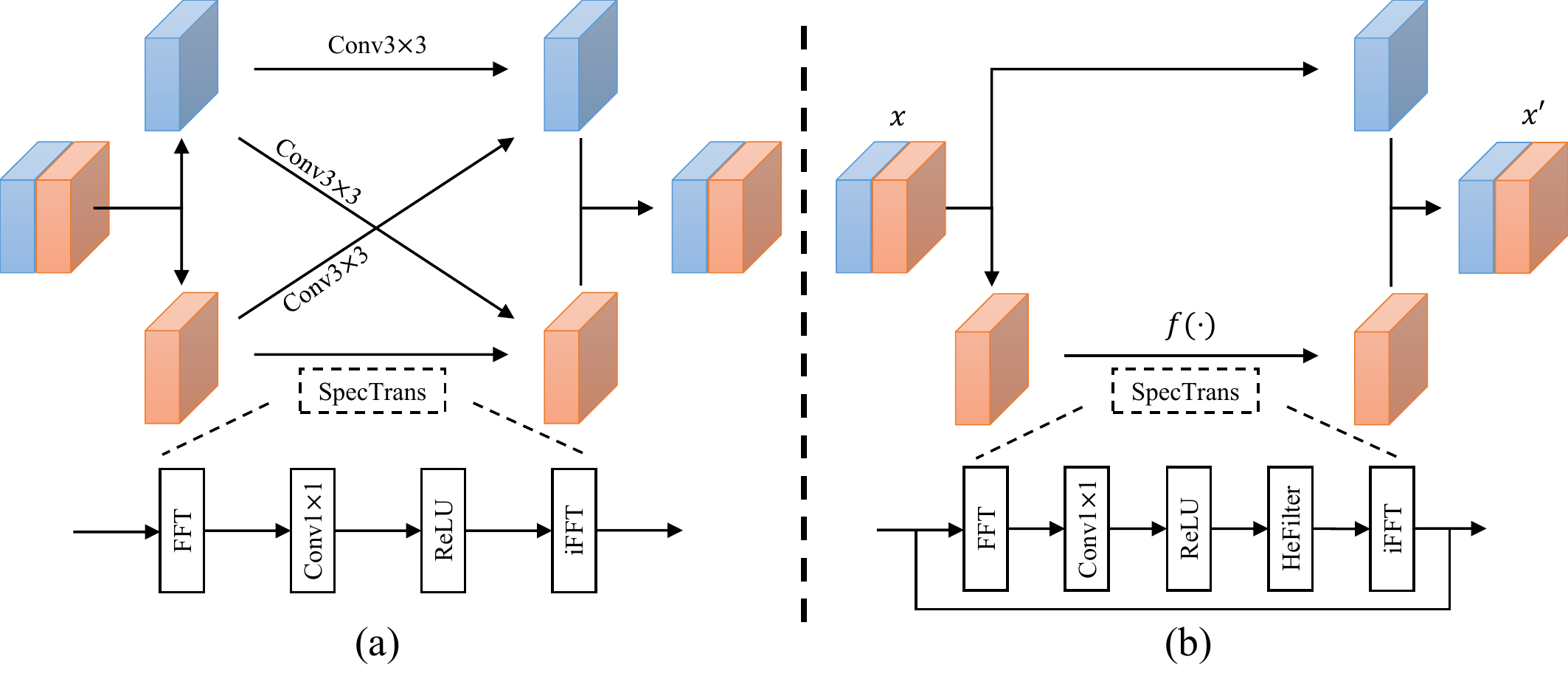}
    \vspace{-0.2cm}
    \caption{This figure shows the structure of FFC~\cite{ffc} (left) and SHU (right). Different from FFC, our SHU does not use any external convolution layers. In the spectral transform, SHU utilizes HeFilter to transform spectral tensors after ReLU, while FFC directly connects ReLU outputs to iFFT. To make things compacted, we do not include Gaussian Split in this figure.
    }
    \vspace{-0.3cm}
    \label{fig:shu}
\end{figure*}

\subsection{Spectral Research in Computer Vision}

Early image spectral research largely focused on low-level vision such as compression~\cite{dct,ft-huang0}, restoration~\cite{wt-huang0}, and denoising~\cite{wt-denoise1,wt-denoise0,ct-denoise0,ft-denoise0}. 
In 1971, Tsai and Huang proposed Transform Coding~\cite{ft-huang0} using the discrete cosine transform, which was later extended into the well-known JPEG format~\cite{dct}. Huang~\cite{wt-huang0} also initiated the pioneering work for image enhancement and restoration using multi-frame discrete Fourier transform (DFT) and inverse-DFT. Popular approaches of image denoising utilize fast Fourier transform (FFT) ~\cite{ft-denoise0} or wavelet transform~\cite{wt-denoise1,wt-denoise0}. In~\cite{ct-denoise0}, Jean \etal proposed two new frequency-domain tools: ridgelet transform and curvelet transform, by which they could recover images from noise with higher perceptual quality than prior works. In recent years, researchers have shown increasing enthusiasm for spectral neural networks. \cite{speccnn} was one of the first works that combined spectral layers with CNN, in which it proposed spectral pooling for dimension reduction. Another work \cite{syncspeccnn} proposed SyncSpecCNN in which a set of 3D features were eigen-decomposed and passed through a CNN for 3D part segmentation. For super-resolution, \cite{wt-sr} decomposed tensors with the wavelet basis and transformed them with fully connected layers. \cite{specbias} explored the inductive bias of CNNs towards low-frequency signals. A similar discovery was mentioned in \cite{ssd-gan}, in which the author performed analysis on the frequency domain of a GAN network. Recently, Chi \etal proposed Fast Fourier Convolution (FFC)~\cite{ffc}, in which tensors were converted between spatial and frequency domain using FFT and inverse-FFT. Chi \etal also showed that FFC could substitute regular Residual Blocks~\cite{resnet} and reach better performance in classification. 

\subsection{Image Completion}

The goal of image completion is to synthesize image content for missing regions. Traditional approaches performed gray-level gradient extension~\cite{inpaint0}, image quilting~\cite{inpaint1}, and patch-based methods \cite{patch-match,criminisi2004region,efros1999texture}. Notwithstanding the success of these approaches in the cases of simple and highly-textured backgrounds, they fail to recover missing semantics and complex structures. Since the popularity of deep learning, \cite{inpaint-dnn0,inpaint-dnn1,inpaint-dnn2} were the first groups to design deep network architectures for inpainting. Satoshi \etal~\cite{inpaint-dnn3} utilized dilated convolutions and adversarial training. ~\cite{inpaint-dnn4} inpainted face images using semantic maps as guidance. 
To address the vanilla convolutions’ drawback of treating all pixels equally, Liu \etal~\cite{partial-conv} introduced partial convolutions in which unknown elements in the input tensor are excluded from the calculations. Yu \etal further improved the performance with contextual attention~\cite{deepfillv1} and gated convolutions~\cite{deepfillv2}. Navasardyan \& Ohanyan~\cite{onionconv} proposed the onion convolutions, in which neighboring patches could be searched and relocated simultaneously with the convolution operation. Sharing a similar spirit with~\cite{patch-match,deepfillv1}, HiFill~\cite{hifill} generated contextual residual to fill in higher resolution textures. Another work, CR-Fill~\cite{crfill} proposed the CR loss, a patch similarity loss designed to reinforce contextual consistency. Zhu \etal~\cite{madf} introduced the MADF module and cascaded refinement decoders. Zeng \etal~\cite{aotgan} introduced the AOT block and utilized soft mask-guided PatchGAN~\cite{patchgan} for the network training. Suvorov \etal introduced LaMa, which is a U-shape structure with FFCs~\cite{ffc}. Zhao \etal~\cite{comodgan} proposed CoModGAN along with the co-modulation idea on top of the StyleGAN~\cite{stylegan-ada,stylegan1,stylegan2}. All these works have achieved plausible results on faces and natural images using free-form masks. 
The concurrent works in diffusion models~\cite{ddpm,ddim} can also be extended to inpainting tasks in which LDM~\cite{ldm} and DALL-E2~\cite{dalle2} shows promising results at a higher computational cost during inference time.

\section{Method}\label{sec:method}
In this section, we illustrate the key designs of our work, including Spectral Hint Unit (SHU), Heterogeneous Filtering Layer (HeFilter), Gaussian Split, and the overall GAN architecture.
The definition of image completion is to restore an RGB image $I_{fake}$ from a masked color image $I=I_{real}\odot M$, where $I_{real}$ is the ground-truth image and $M$ is the mask, in which the known pixels have value 1 and the unknown pixels are represented with zeros.

\begin{figure*}[t]
\centering
    \hspace{0.05\textwidth}
    \begin{subfigure}[b]{0.25\textwidth}
    \centering
        \includegraphics[width=\textwidth]{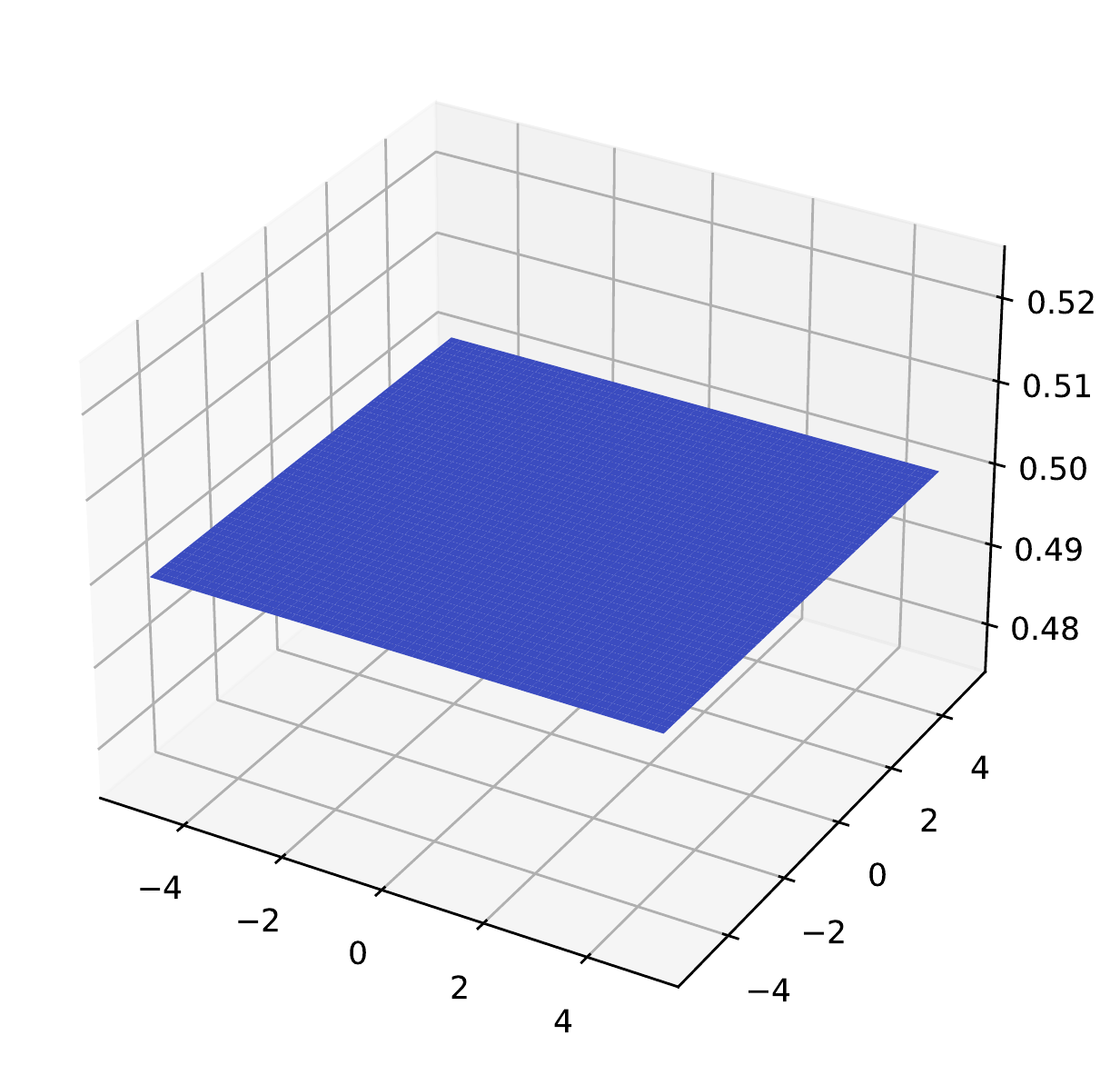}
    \subcaption{Conv1$\times$1}
    \end{subfigure}
    \hfill
    \begin{subfigure}[b]{0.25\textwidth}
    \centering
        \includegraphics[width=\textwidth]{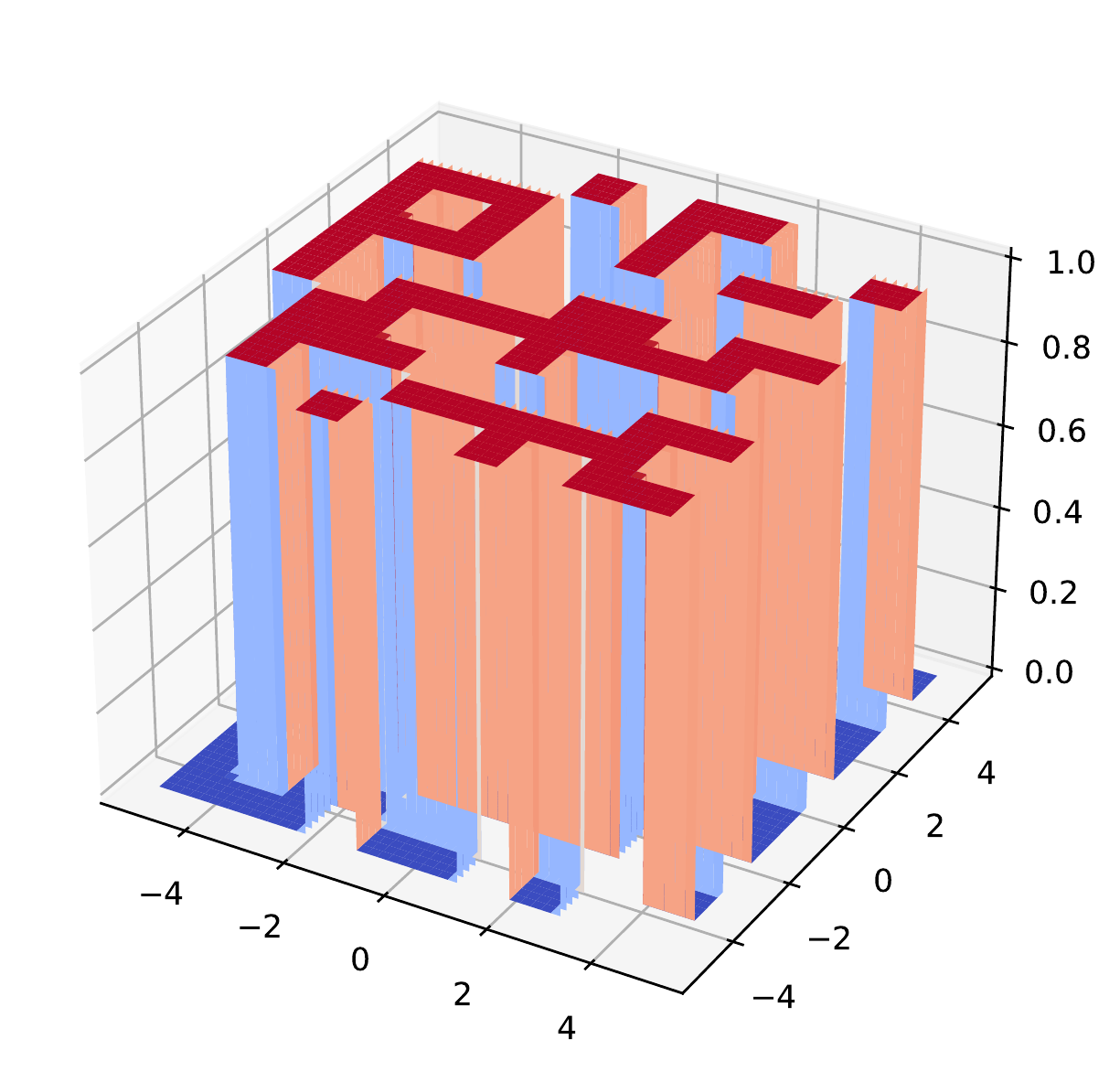}
    \subcaption{ReLU}
    \end{subfigure}
    \hfill
    \begin{subfigure}[b]{0.25\textwidth}
    \centering
        \includegraphics[width=\textwidth]{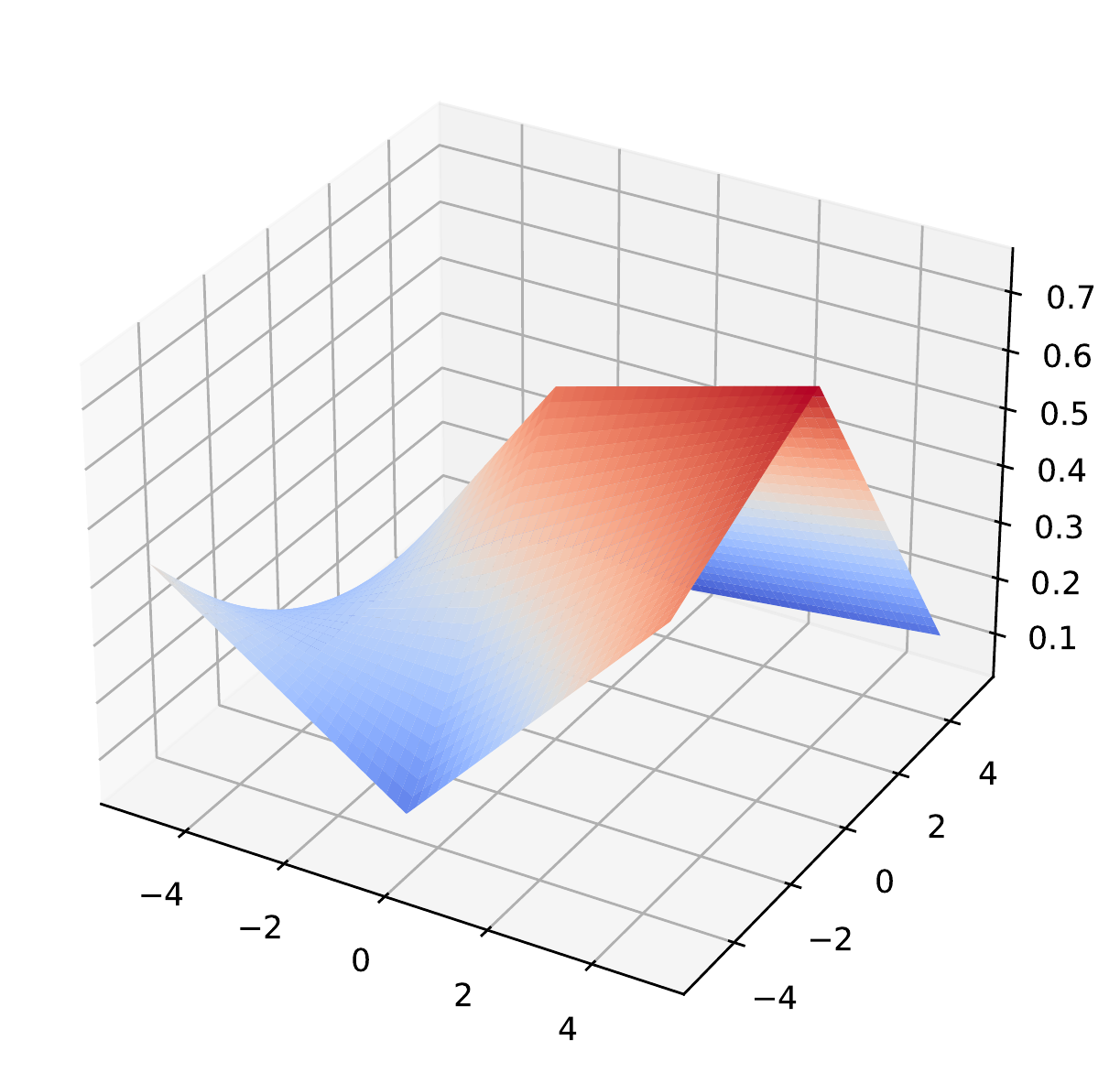}
    \subcaption{HeFilter}
    \end{subfigure}
    \hspace{0.05\textwidth}
\caption{Graphic explanations on manipulating spectral tensors using different types of layers. Conv1$\times$1 is a homogeneous operation that treats the complex vector-space equally over the 2D frequency domain. ReLU works as a band-pass filter by changing negative components to zeros. Our HeFilter applies a smooth varying mapping function on the frequency domain and manipulates complex vectors based on their spectral location.}
\vspace{-0.3cm}
\label{fig:hf_explain}
\end{figure*}

\begin{figure*}[t]
\centering
    \includegraphics[width=0.96\textwidth]{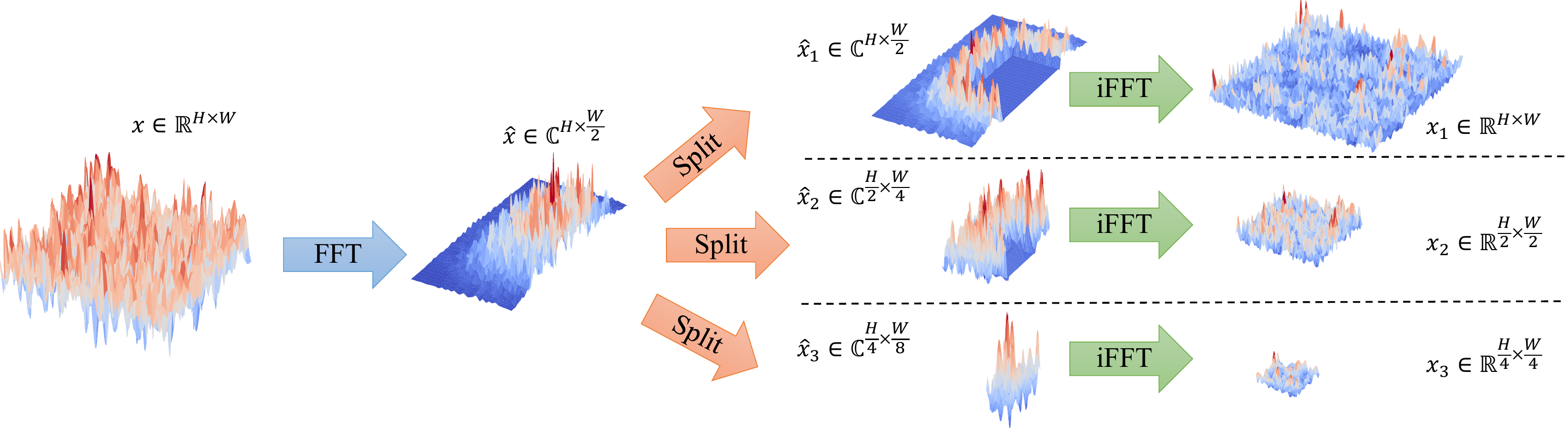}
    \vspace{-0.2cm}
    \caption{A graphic illustration on the Vanilla Split. Our Gaussian Split is an extended version adding a Gaussian mask on each split level for better anti-aliasing. Due to the limited space, we only show the Vanilla Split here.}
    \vspace{-0.2cm}
\label{fig:split_explain}
\end{figure*}

\subsection{Spectral Transform with SHU}

Spectral Hint Unit (SHU) transforms tensors using a neat FFT$-$Network$-$iFFT pipeline (see Fig. \ref{fig:shu}). The recent work FFC~\cite{ffc} also suggested a similar structure in which the author blended spectral transform inside a densely connected convolutional network. Different from FFC, SHU is light-weighted because it has no extra convolutions outside the spectral transform. More precisely, let $x\in \mathbb{R}^{N\times H\times W}$ be an $N$-channeled tensor with height and width equal to $H$ and $W$ respectively. Then the output of SHU is the tensor $x'$ with the same dimentionality formed by the following way:

\vspace{-0.5cm}
$$
x' = \text{concat}\left(x_{[0...N-K]}, \hspace{0.3em} x_{[N-K...N]} + f\left(x_{[N-K...N]}\right)\right)
$$
$$
f = \text{iFFT} \circ g \circ \text{FFT}
$$
$$
g = \text{HeFilter} \circ \text{ReLU} \circ \text{Conv}1\times1
$$

\noindent Our design fits GAN training for the following reasons: 
a) local operations such as a convolution should be well-handled by the synthesis network with modulated convolutions; 
b) GAN training should maintain a subtle balance between the spectral and spatial transforms, and it should not overwhelm tensors with spectral information.  

Like the spectral transform of FFC~\cite{ffc}, Conv1$\times$1 maps between two complex vector-spaces uniformly in the frequency domain.
ReLU is the non-linearity operation that filters out all negative components in the vector-space. Lastly, HeFilter performs heterogeneous filtering in which the mapping is a smooth-varying function over the 2D spectral location.
We will describe HeFilter with more details in the next subsection. 
In summary, the spectral transform of SHU is as follows:

\vspace{-0.2cm}
\begin{enumerate}[label=\alph*),leftmargin=0.3in]
    \setlength\itemsep{-0.2em}
    \item $\mathbb{R}^{K \times H \times W} \rightarrow \mathbb{C}^{K \times H \times \frac{W}{2}}$ (FFT) 
    \item $\mathbb{C}^{K \times H \times \frac{W}{2}} \rightarrow \mathbb{C}^{K \times H \times \frac{W}{2}}$ (HeFilter $\circ$ ReLU $\circ$ Conv)
    \item $\mathbb{C}^{K \times H \times \frac{W}{2}} \rightarrow \mathbb{R}^{K \times H \times W}$ (iFFT)
\end{enumerate}
\vspace{-0.4cm}


\begin{figure*}[t]
    \centering
    \vspace{0.2cm}
    \includegraphics[width=0.97\textwidth]{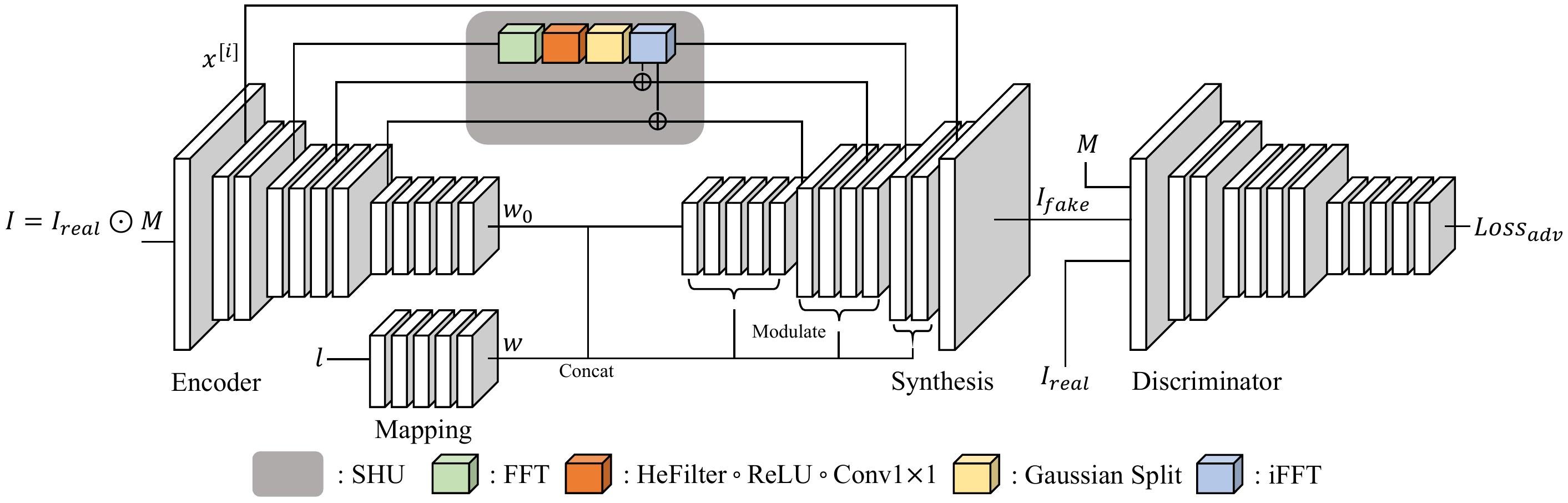}
    \vspace{-0.2cm}
    \caption{This figure shows the overall structure of our SH-GAN in which our SHU is highlighted in the gray area. }
    \vspace{-0.1cm}
    \label{fig:network}
\end{figure*}

\subsection{Heterogeneous Filtering and Gaussian Split}

As mentioned earlier, one of the contributions of this work is to introduce two novel spectral processing strategies: Heterogeneous Filtering and Gaussian Split, that well-fit the deep learning training scheme. 


\textbf{Heterogeneous Filtering.} Recall that FFC~\cite{ffc} transforms spectral tensors with Conv1$\times$1 and ReLU, which can also be viewed as a homogeneous operation and a band-pass filter. 
In many cases, ReLU is a necessary step but not a recommended end operation for spectral transform with the following reasons: a) it deactivates a frequency band with no recovery; b) it introduces aliasing effects due to non-smoothness; and c) it responses according to magnitude instead of location (\ie requency band).
Therefore, we create HeFilter, inside of which a heterogeneous filtering strategy is introduced transforming the complex vector-space via a smooth-varying function over the frequency domain.
More precisely, HeFilter learns several weight matrices scattered on an even-spaced 2D frequency domain.
During propagation, HeFilter linearly interpolates these weights and multiplies them with the corresponding spectral vector. Figure~\ref{fig:hf_explain} explains the characteristic of Conv1$\times$1, ReLU, and HeFilter in the spectral transform. We prepare a total of $3\times 2$ weights on the 2D spectral domain. The asymmetry along each dimension is because the FFT of $\mathbb{R}^{K \times H \times W}$ is $\mathbb{C}^{K \times H \times \frac{W}{2}}$, and the skipped half is the reflected complex conjugate. We do not impose constraints on learning these weights, thus HeFilter can be low-pass, band-pass or high-pass depending on the weights it learns. This also explains the name \textit{heterogeneous} because it only guarantees the discrepancy of the transform on various spectral locations. A simple extension of our HeFilter is to use a larger grid size such as $5\times 3$ or $7\times 4$. Nevertheless, we notice sufficient improvements on model performance using grid size $3\times 2$. HeFilter with larger grid size is left for further studies. 

\textbf{Gaussian Split.} Popular generative models~\cite{progressive-gan,stylegan-ada,stylegan3,stylegan1,stylegan2} adopt the progressive structure in which low-resolution features are gradually elaborated into high-resolution features. We fit our SHU in this structure by adding a unique Gaussian Split, segregating spectral tensors into multi-resolution sub-tensors before the iFFT operation.
A fundamental property of the Fourier Transform $\mathcal{F}$ is linearity, in which the FT of the addition of the functions $f_1$ and $f_2$ equals the addition of the individual FTs of $f_1$ and $f_2$ (see Eq.~\ref{eq:fft_linearity}).

\vspace{-0.2cm}
\begin{equation}
    \begin{gathered}
    f(x) \xleftrightarrow{\mathcal{F}} \hat{f}(\omega) 
    \\ 
    \alpha f_1(x) + \beta f_2(x) \xleftrightarrow{\mathcal{F}} 
    \alpha \hat{f_1}(\omega) + \beta \hat{f_2}(\omega) 
    \end{gathered}
\label{eq:fft_linearity}
\end{equation}

\noindent Utilizing the property mentioned above, we can split any spectral signal $\hat{x} = \mathcal{F}(x)$ into several sub-signals $\hat{x}_i, i\in \{1 \dots n\}$. As long as $\sum_{i}{\hat{x}_i} = \hat{x}$, we expect no information loss on $x=\sum_{i}{\mathcal{F}^{-1}(\hat{x}_i)}$, and this property holds for FFT on discrete tensors. For convenience, we use the same set of symbols $\hat{x},\hat{x}_i$ representing spectral tensors. The graphic explanation of the split is highlighted in Figure~\ref{fig:split_explain}, from which one may notice that our decomposition, like the Wavelet Transform, automatically segregates signals based on their frequency bands. For example, a two-level Vanilla Split is formulated in Eq.~\ref{eq:split} and~\ref{eq:vanilla_split}, in which we migrate all low frequency values from the large tensor to the small one. 

\begin{equation}
    \hat{x} \in \mathbb{C}^{H \times \frac{W}{2}}
    \xleftrightarrow{\text{split}} 
    \left(
    \hat{x_1} \in \mathbb{C}^{H \times \frac{W}{2}}, 
    \hat{x_2} \in \mathbb{C}^{\frac{H}{2} \times \frac{W}{4}}
    \right)
\label{eq:split}
\end{equation}

\vspace{-0.3cm}

\begin{equation}
    \begin{gathered}
        \hat{x_1}_{[i,j]} =
        \begin{cases}
            \hat{x}_{[i,j]} & (i,j)\notin(\frac{H}{4}...\frac{3H}{4}, 0...\frac{W}{4})\\
            0 & (i,j)\in(\frac{H}{4}...\frac{3H}{4}, 0...\frac{W}{4})
        \end{cases}
        \\
        \hat{x_2} = \hat{x}_{[\frac{H}{4}...\frac{3H}{4}, 0...\frac{W}{4}]}
    \end{gathered}
    \quad
    \text{\footnotesize(Vanilla)}
\label{eq:vanilla_split}
\end{equation}

\noindent Our Gaussian Split is an upgraded version of Vanilla Split, in which we smooth each splits with Gaussian weight maps $\mathcal{N}$ to minimize the aliasing effect. (see Eq.~\ref{eq:gaussian_split}). The center of $\mathcal{N}$ is at $(\frac{H}{2}, 0)$ and the standard deviation $\sigma$ is proportional to the corresponding resolution. 
Since the Fourier transform of a Gaussian function is another Gaussian function, applying Gaussian maps in the frequency domain is equivalent to applying a Gaussian blur filter on the spatial domain that we usually do before downsampling. One may also notice that the multi-level Gaussian Split serves as the well-known Difference of Gaussian (DoG)~\cite{dog} in the spatial domain. We will demonstrate the effectiveness of our Heterogeneous Filtering and Gaussian Split in section~\ref{sec:experiments}. 

\vspace{-0.3cm}

\begin{equation}
    \begin{gathered}
        \hat{x_1}_{[i,j]} =
        \begin{cases}
            \hat{x}_{[i,j]}
            & (i,j)\notin(\frac{H}{4}...\frac{3H}{4}, 0...\frac{W}{4})\\
            \hat{x}_{[i,j]} \times (1-\mathcal{N}_{[i,j]}) 
            & (i,j)\in(\frac{H}{4}...\frac{3H}{4}, 0...\frac{W}{4})
        \end{cases}
        \\
        \hat{x_2} = \hat{x}_{[\frac{H}{4}...\frac{3H}{4}, 0...\frac{W}{4}]} 
        \times \mathcal{N}_{[\frac{H}{4}...\frac{3H}{4}, 0...\frac{W}{4}]}
    \\
    \quad\text{\footnotesize(Gaussian)}
    \end{gathered}
\label{eq:gaussian_split}
\end{equation}

\begin{table*}
\centering
\resizebox{\textwidth}{!}{
    \begin{tabular}{
            L{3.5cm}
            C{1.5cm}
            C{1.5cm}
            C{1.5cm}
            C{1.5cm}
            C{1.5cm}
            C{1.5cm}
            C{1.5cm}
            C{1.5cm}}
        \toprule
            & \multicolumn{4}{c}{\textbf{FFHQ 256}}
            & \multicolumn{4}{c}{\textbf{Places2 256}}
            \\
        \cmidrule(l{0.5em}r{0.5em}){2-5}
        \cmidrule(l{0.5em}r{0.5em}){6-9}
            \textbf{Method}
            & FID($\downarrow$)
            & LPIPS($\downarrow$) 
            & PSNR($\uparrow$)
            & SSIM($\uparrow$)
            & FID($\downarrow$)
            & LPIPS($\downarrow$)
            & PSNR($\uparrow$)
            & SSIM($\uparrow$)
            \\
        \midrule
        CoModGan (small)
            & 5.0184
            & 0.2579
            & 16.31
            & 0.5892
            & 9.5159
            & 0.3995
            & 14.49
            & 0.4914
            \\
        CoModGan (official)
            & 4.7755
            & 0.2568
            & 16.24
            & 0.5913
            & 9.3621
            & 0.3990
            & 14.50
            & 0.4923
            \\
        LaMa
            & 32.7035
            & 0.2590
            & \best{17.58}
            & 0.6277
            & 23.7409
            & \best{0.3679}
            & 16.58
            & \best{0.5448}
            \\
        DeepFillV2
            & 50.9323
            & 0.3204
            & 16.11
            & 0.5569
            & 46.2012
            & 0.4166
            & 14.97
            & 0.4913
            \\
        CR-Fill
            & -
            & -
            & - 
            & -
            & 40.9690
            & 0.3957
            & 15.28
            & 0.4925
            \\
        Onion-Conv
            & -
            & -
            & -
            & -
            & 42.4625
            & 0.4360
            & 15.03
            & 0.5046
            \\
        MADF
            & 33.6207
            & 0.2800
            & 17.54
            & \best{0.6279}
            & 66.2659
            & 0.3889
            & \best{16.61}
            & 0.5360
            \\
        AOT-GAN
            & 73.7962
            & 0.4270
            & 15.60
            & 0.5956
            & 90.6184
            & 0.5139
            & 14.87
            & 0.4790
            \\
        \midrule
        (ours - small)
            & 4.8225
            & 0.2558
            & 16.36
            & 0.5891
            & 8.3078
            & 0.3969
            & 14.50
            & 0.4918
            \\
        (ours - regular)
            & \best{4.3459}
            & \best{0.2542}
            & 16.37
            & 0.5911
            & \best{7.5036}
            & 0.3940
            & 14.58
            & 0.4958
            \\
        \bottomrule
    \end{tabular}
}
\vspace{0.1cm}
\caption{
    This table compares the performance of prior models with our SH-GAN on datasets FFHQ and Places2 under resolution 256.
}
\label{table:256_performance}
\end{table*}

\begin{table*}
\centering
\resizebox{\textwidth}{!}{
    \begin{tabular}{
            L{3.5cm}
            C{1.5cm}
            C{1.5cm}
            C{1.5cm}
            C{1.5cm}
            C{1.5cm}
            C{1.5cm}
            C{1.5cm}
            C{1.5cm}}
        \toprule
            & \multicolumn{4}{c}{\textbf{FFHQ 512}}
            & \multicolumn{4}{c}{\textbf{Places2 512}}
            \\
        \cmidrule(l{0.5em}r{0.5em}){2-5}
        \cmidrule(l{0.5em}r{0.5em}){6-9}
            \textbf{Method}
            & FID($\downarrow$)
            & LPIPS($\downarrow$) 
            & PSNR($\uparrow$)
            & SSIM($\uparrow$)
            & FID($\downarrow$)
            & LPIPS($\downarrow$)
            & PSNR($\uparrow$)
            & SSIM($\uparrow$)
            \\
        \midrule
        CoModGan (small)
            & 3.9420
            & 0.2497
            & 18.38
            & 0.6911
            & 8.8390
            & 0.3464
            & 15.96
            & 0.5925
            \\
        CoModGan (official)
            & 3.6996
            & 0.2469
            & 18.46
            & 0.6956
            & 7.9735
            & 0.3420
            & 16.00
            & 0.5953
            \\
        LaMa
            & 19.5577
            & 0.2871
            & 18.99
            & \best{0.7178}
            & 12.6721
            & \best{0.3158}
            & 17.12
            & \best{0.6521}
            \\
        DeepFillV2
            & 32.8696
            & 0.3283
            & 18.29
            & 0.6886
            & 29.7345
            & 0.3802
            & 15.91
            & 0.5953
            \\
        CR-Fill
            & -
            & -
            & -
            & -
            & 26.6398
            & 0.3593
            & 16.52
            & 0.6038
            \\
        Onion-Conv
            & -
            & -
            & -
            & -
            & 25.7480
            & 0.3999
            & 15.02
            & 0.6061
            \\
        MADF
            & 17.1962
            & 0.2688
            & \best{19.62}
            & 0.7196
            & 29.4928
            & 0.3299
            & \best{17.77}
            & 0.6239
            \\
        AOT-GAN
            & 36.1344
            & 0.3403
            & 18.14
            & 0.7131
            & 46.7640
            & 0.3976
            & 16.84
            & 0.6029
            \\
        \midrule
        (ours - small)
            & 3.7460
            & 0.2491
            & 18.36
            & 0.6897
            & 7.6122
            & 0.3455
            & 15.95
            & 0.5926
            \\
        (ours - regular)
            & \best{3.4134}
            & \best{0.2447}
            & 18.43
            & 0.6936
            & \best{7.0277}
            & 0.3386
            & 16.03
            & 0.5973
            \\
        \bottomrule
    \end{tabular}
}
\vspace{0.1cm}
\caption{
    This table compares the performance of prior models with our SH-GAN on datasets FFHQ and Places2 under resolution 512.
}
\vspace{-0.4cm}
\label{table:512_performance}
\end{table*}

\subsection{Network Architecture}

Similar to CoModGAN~\cite{comodgan}, our model is a U-shape architecture containing an encoder and a synthesis network. As Figure~\ref{fig:network} shows, we first pass the masked input image $I$ into the encoder, inside which $I$ is encoded into a set of feature maps $x^{[i]}$ with resolution $i$, and a global vector $w_0$. 
We then split $K$ channels from $x^{[i]}$ ($K=32, i=64$), and pass it to SHU for spectral transform. 
Inside SHU, we create a series of wavelet-like feature maps $x_{i}, i\in\{4, ..., 64\}$ using Gaussian Split, in which the low-frequency information is encoded in the low-resolution feature maps (\eg $x_{4}$) and the high-frequency information is encoded in the high-resolution feature maps (\eg $x_{64}$). We then add back $x_{i}$ to the corresponding $x^{[i]}$. For $i>64$, we directly pass $x^{[i]}$ to synthesis blocks. Recall that we adopt StyleGAN2~\cite{stylegan-ada} as our synthesis network. We then modulate the synthesis network using the concatenation of $w$ and $w_0$, in which $w$ is the projected vector of the latent code $l$ using the mapping network, and connecting $x^{[i]}$ with each synthesis block via addition. 

During the training time, we use a StyleGAN2 discriminator for our adversarial loss. We also use path length regularization on the generator with $w_{pl}=2$, and $R_1$ regularization on the discriminator with $\gamma=10$. Other training details can be found in Section~\ref{sec:training_details}.

\subsection{Mask generation}\label{sec:mask_generation}

We use the same free-form mask generation algorithm as DeepFillV2~\cite{deepfillv2} and CoModGAN~\cite{comodgan}. These masks are drawn using multiple brush strokes and rectangles. The width of the brush stroke is sampled from $U(12,48)$ and the number of strokes is chosen  randomly from $U(0,20)$, where $U(\cdot)$ represents the discrete uniform distribution. Meanwhile, we sample $U(0,5)$ rectangles up to the full size of the input image, and $U(0,10)$ rectangles up to the half-size. For more details, please see Appendix~\ref{appendix:masks}.

\section{Experiments}\label{sec:experiments}
This section goes through the details about our dataset, metrics, settings, experiments, and other studies. We provide comprehensive comparisons between our SH-GAN and other prior works through scores and images. 

\begin{figure*}[h]
    \centering
    \includegraphics[width=0.9\textwidth]{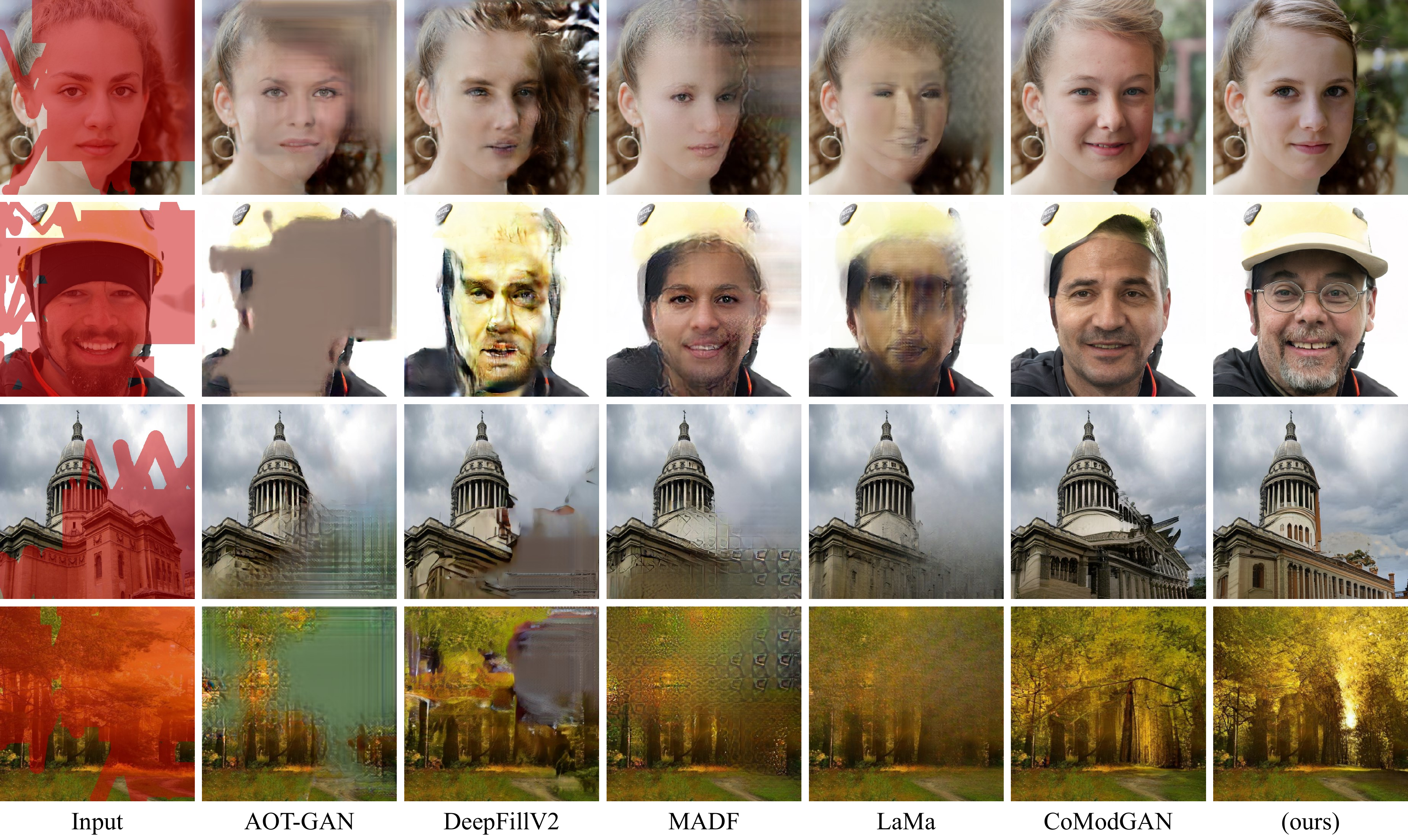}
    \caption{The qualitative comparison between prior approaches and our approach using free-form masks. For other types of masks, please see Appendix~\ref{appendix:extra_results}.}
    \label{fig:qcompare}
    \vspace{-0.4cm}
\end{figure*}

\subsection{Datasets and Metrics}

We use three datasets: FFHQ~\cite{stylegan1}, Places2~\cite{places2}, and DTD~\cite{dtd}. FFHQ contains 70,000 high-resolution well-aligned face images, in which we split out 60,000 images for training and use the remaining 10,000 images for validation. Places2 contains 8,026,628 images in its training set and 36,500 images in its validation set. The contents of Places2 are regular scenes and objects. We maintain the original train-validation split for our experiments. DTD contains 5,640 categorized texture images, among which 3,760 are from the train and validation set, and 1,880 are from the test set. We train our models using the train and validation sets and evaluate them using the test set.

We use \textit{Fréchet Inception Distance} (FID)~\cite{fid} as our primary metric. FID is a statistical score that calculates the distance between the distributions of real and synthetic features. Besides, we also adopt \textit{Learned Perceptual Image Patch Similarity} (LPIPS)~\cite{lpips}, \textit{Peak Signal-to-Noise Ratio} (PSNR), and \textit{Structural Similarity Index} (SSIM)~\cite{ssim} to gauge models from different angles. We will show all the metric scores in Section~\ref{sec:result_comparison}.

\vspace{-0.1cm}
\subsection{Training Details}\label{sec:training_details}

Many of the training settings of this work closely follow CoModGAN~\cite{comodgan} and StyleGAN2~\cite{stylegan-ada}. We use Adam~\cite{adam} optimizer with $\beta=(0, 0.99)$ for our generator and discriminator. The learning rates are 0.001 for FFHQ/Places2, and 0.002 for DTD. Besides, the training lengths are 25 million images on FFHQ, 50 million on Places2, and 10 million on DTD. We apply both path length regularization and $R_1$ regularization during the training, in which we set $w_{pl}=2$ and $\gamma=10$. The batch size is 32 for all models on all datasets. Like StyleGAN2, we compute the exponential moving average of our generator with momentum 0.99993 (\ie half-decay at 20,000 images with batch size 32). As shown in Tables~\ref{table:256_performance} and~\ref{table:512_performance}, we evaluate SH-GAN with small and standard settings under resolutions 256 and 512. SH-GAN (small) is a reduced version of SH-GAN with base channel decreased from 32,768 to 16,384 (See Appendix~\ref{appendix:architecture}). Such small and standard settings match the small and official versions of CoModGAN in terms of the model size. We train the small model with 4 GPUs and the standard model with 8 GPUs. Besides, we use 2080Ti for resolution 256 and A100 for resolution 512. 

\begin{figure}[t!]
    \centering
    \includegraphics[width=\columnwidth]{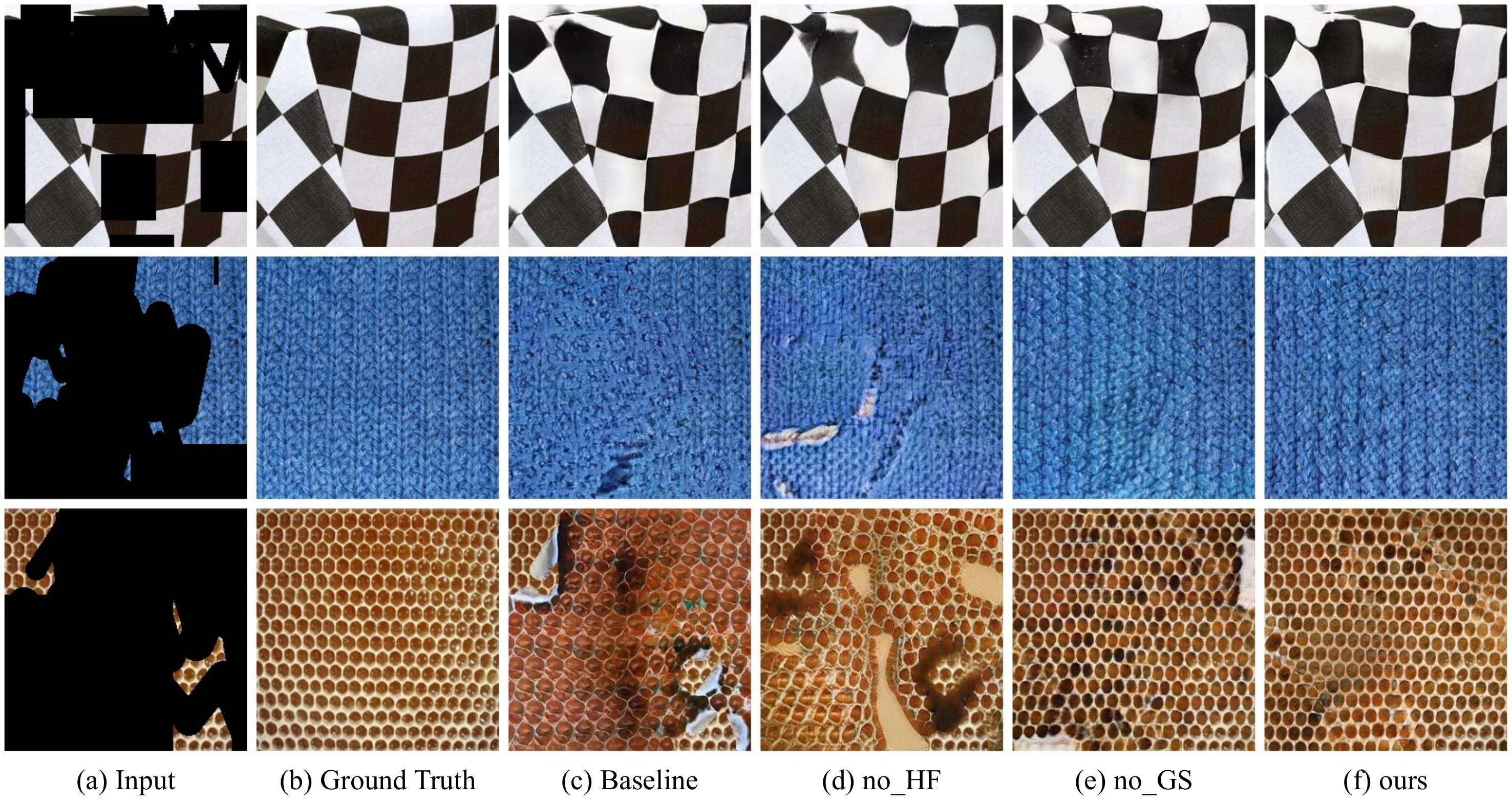}
    \vspace{-0.6cm}
    \caption{This figure shows the qualitative comparison of various settings in our ablation study. The performance is gradually improved with our SHU and our spectral processing strategies.}
    \vspace{-0.3cm}
    \label{fig:dtd_aba}
\end{figure}

\begin{figure*}[t!]
    \centering
    \vspace{0.1cm}
    \includegraphics[width=\textwidth]{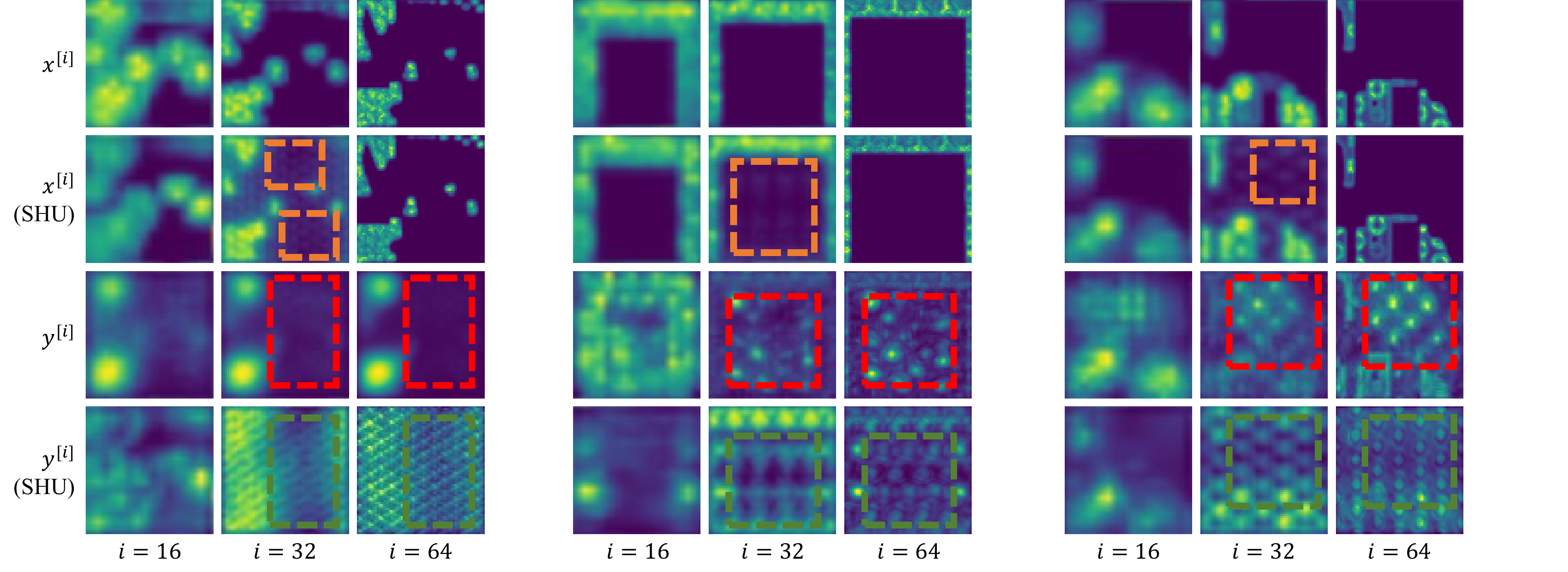}
    \caption{This figure shows three examples comparing intermediate features with SHU and without SHU. The {\color{orange}orange} dashed boxes highlight the spectral hints added by SHU. The {\color{olive}green} and  {\color{red}red} dashed boxes compare the locations where features are generated with and without fidelity due to the spectral hints.}
    \vspace{-0.2cm}
    \label{fig:dtd_abb}
\end{figure*}

\subsection{Results Comparison}\label{sec:result_comparison}

Tables~\ref{table:256_performance} and~\ref{table:512_performance} compare the performances of SH-GAN with other prior works~\cite{lama,deepfillv2,aotgan,crfill,comodgan,madf} on FFHQ and Places2. As mentioned in Section~\ref{sec:mask_generation}, we adopt the free-from masks originated from the CoModGAN paper. Among the four metrics, FID/LPIPS gauge perception quality, and PSNR/SSIM gauge pixel accuracy. 
Note that these metrics reveal image quality in a different way so they may disagree in numbers. For most of the prior works, except for CoModGAN, we have downloaded the official code and models for evaluation. 
We re-implemented CoModGAN in Pytorch, and trained and tested the replicated version. The FID scores on the replicated CoModGAN match the FID scores in the original paper. As a result, SH-GAN reaches 4.3459 and 7.5036 on FFHQ and Places2 datasets respectively for the resolution 256, and 3.5713 and 7.8482 for the resolution 512. SH-GAN surpasses all other approaches in terms of FID and becomes the new state-of-the-art.

In addition to the free-form mask experiments, we also tried other types of masks such as the LaMa-style~\cite{lama} narrow and wide masks. These detailed performance can be found in the Appendix~\ref{appendix:extra_results}. 

\subsection{Extended Studies}

For this section, we carry out extended experiments to justify our new designs (\ie SHU, Heterogeneous Filtering, and Gaussian Split) over the prior challenges such as pattern unawareness, blurry texture, and structure distortion. 

Our first experiment is an ablation study using the DTD dataset, in which we trained several models and excluded SHU, Heterogeneous Filter, or Gaussian Split, respectively. We focused on the DTD dataset because texture images are highly structured images that could make obvious cases for comparison. In Table~\ref{table:aba}, we show that the full version of our model performed FID 48.58, lower than the model without Gaussian Split (\ie ours - no\_GS) by 1.48, lower than the model without Heterogenerous Filtering (\ie ours - no\_HF) by 1.83, and lower than the baseline (\ie CoModGAN~\cite{comodgan}) by 3.35. The LPIPS, PSNR and SSIM scores agree with our FID score. Moreover, we clearly show in Figure~\ref{fig:dtd_aba} that our model generates sharp and robust patterns even when the mask coverage is large. Our HeFilter is very helpful in cases with complex structures and our Gaussian Split helps to remove the aliasing effect to make the pattern sharper. 

\begin{table}[t!]
\centering
\resizebox{\columnwidth}{!}{
    \begin{tabular}{
            L{2cm}
            |C{0.75cm}
            C{0.5cm}
            C{0.5cm}
            |C{1.5cm}
            C{1.5cm}}
        \toprule
            \textbf{Models}
            & SHU
            & HF
            & GS
            & FID($\downarrow$)
            & LPIPS($\downarrow$)
            \\
        \midrule
        baseline
            &
            &
            &
            & 51.9289
            & 0.3628
            \\
        ours - no\_HF
            & \checkmark
            & 
            & \checkmark
            & 50.4074
            & 0.3614
            \\
        ours - no\_GS
            & \checkmark
            & \checkmark
            &
            & 50.0634
            & 0.3573
            \\
        ours
            & \checkmark
            & \checkmark
            & \checkmark
            & \textbf{48.5814}
            & \textbf{0.3519}
            \\
        \bottomrule
    \end{tabular}
}
\vspace{0.1cm}
\caption{
    The ablation study on DTD~\cite{dtd} with SHU, Heterogeneous Filtering (HF), and Gaussian Split (GS) removed respectively.
}
\vspace{-0.4cm}
\label{table:aba}
\end{table}

In our second experiment, we extract the skip feature maps $x^{[i]}$ generated by the encoder, and the intermediate feature maps $y^{[i]}$ generated by the synthesis blocks, $i \in \{16, 32, 64\}$. We then compute the 2-norm of each feature map along the channel axis. To make clear comparisons, our SHU is only connected on resolution 32 without splitting. Figure~\ref{fig:dtd_abb} shows the impact of SHU over these features, in which readers may notice that SHU provides critical hints on patterns to its downstream synthesis blocks for texture generation.






\section{Conclusions}
We introduce SH-GAN, a novel image completion approach that transforms deep features with spectral hints in the modulated GAN framework. Throughout this paper, we reveal the details of our newly designed module: SHU, and introduce our new spectral transform strategies: Heterogeneous Filtering and Gaussian Split. With inclusive experiments, we show that all our designs are very useful in solving challenging inpainting cases with large-scale free-form missing regions. We believe that our SHU and spectral transform strategies are worth exploring further in other computer vision tasks.

{\small
\bibliographystyle{ieee_fullname}
\bibliography{egbib}
}

\clearpage

\onecolumn
\begin{appendices}

\vspace{0.5cm}

\section{Architecture}\label{appendix:architecture}
In the main article, we have introduced two model settings, small and standard, and the sole difference between our small and standard settings is to shrink the base channel number by half. The base channel number is a hyper-parameter that predominantly affects memory usage of the model during training and inference. 
In each encoder and synthesis block, the channel number $N$ of its convolution layers is computed by an equation connecting the base channel number $N_{base}$, the spatial resolution $N_{res}$ and a pre-defined max channel $N_{max}$ (see Eq.~\ref{eq:1}). For our small model, $N_{base}$ is set to 16,384. For our standard model, $N_{base}$ is set to 32,768. All models use $N_{max}$ equals to 512. 

\begin{equation}
    N = \min\left(\frac{N_{base}}{N_{res}}, N_{max}\right)
\label{eq:1}
\end{equation}

\noindent Our experiments showed that the standard model (\ie $N_{base}=32768$) might make the GAN training unstable using batch size 32 on 4 GPUs. Researchers were suggested to use 8 GPUs to make the batch size per GPU no larger than 4 when training the standard model. Nevertheless, our small model was more friendly towards 4 GPU training, helping run the experiment with less computational resources. The GPU memory usage on our small and standard model on resolution 256 and 512 experiments were approximately 10G and 18G per GPU. Remember that our small model could fit batch size 8 per GPU, two times larger than the standard model. On the other hand, the model size in terms of parameters does not significantly differ between our small and regular settings. For resolution 512, the small model contains 68.2 million parameters while the standard model contains 79.8 million parameters (\ie 17\% more parameters). The model's sizes are 68.0 million and 79.2 million parameters for resolution 256, respectively. In summary, we list out the detailed architecture in Table~\ref{table:architecture} for better illustrations.

\section{Evaluation Details}\label{appendix:evaluation_details}
SH-GAN has been fully implemented in PyTorch. Simultaneously, we replicated CoModGAN~\cite{comodgan} using PyTorch apart from its original TensorFlow implementation. When we evaluated SH-GAN and CoModGAN, we followed the common five-run rules, and we took the mean values for all scores (\ie FID, LPIPS, PSNR, and SSIM). The variations of the results came from three places: a) the latent codes were randomly sampled from the normal distribution $\mathcal{N}(0,1)$; b) random noises with learned magnitude were injected into each synthesis block; c) masks were generated randomly. We noticed a typical $\pm$0.3 on FID score around the mean, which aligned with our expectation.

For other models, we used our mask generation rules to create several sets of fixed masks and then evaluated these models using the official demo code provided in their Github. LaMa~\cite{lama} and Onion-Conv~\cite{onionconv} mentioned in their work that they applied the same model for all resolutions; thus, we followed their evaluation scheme. For DeepFillV2~\cite{deepfillv2} and CR-Fill~\cite{crfill}, we downloaded separate models for resolutions 256 and 512; thus, we evaluated using these models correspondingly. For MADF~\cite{madf} and AOT-GAN~\cite{aotgan}, they only provided resolution 512 models. Therefore, in our resolution 256 evaluations, we upsampled both images and masks into 512$\times$512, executed the model, and then downsampled the output images back to 256$\times$256. CR-Fill~\cite{crfill} and Onion-Conv~\cite{onionconv} didn't train on face datasets, so we skipped those experiments.


\begin{table*}[t!]
\centering
\resizebox{\textwidth}{!}{
    \begin{tabular}{
            L{0.5cm}
            |C{1.5cm}
            |L{3.8cm}
            |L{3.8cm}
            |L{3.8cm}
            |L{3.8cm}}
        \hline
            & 
            \thead{block \\ resolution} &
            \thead{model 256 \\ small} &
            \thead{model 256 \\ standard} &
            \thead{model 512 \\ small} &
            \thead{model 512 \\ standard} \\
        \hline
            \parbox[t]{2mm}{\multirow{19}{*}{\rotatebox[origin=c]{90}{encoder}}} &
            \multirow{3}{*}{$512 \times 512$} &
            & 
            & 
            conv1$\times$1 ($4 \rightarrow 32$ ) &
            conv1$\times$1 ($4 \rightarrow 64$) \\

            && 
            & 
            & 
            conv3$\times$3 ($32 \rightarrow 32$) &
            conv3$\times$3 ($64 \rightarrow 64$) \\

            && 
            & 
            &
            conv3$\times$3 ($32 \rightarrow 64$, ds) &
            conv3$\times$3 ($64 \rightarrow 128$, ds) \\

        \cline{2-6}
            & \multirow{3}{*}{$256 \times 256$} &
            conv1$\times$1 ($4 \rightarrow 64$) &
            conv1$\times$1 ($4 \rightarrow 128$) &
            &
            \\

            &&
            conv3$\times$3 ($64 \rightarrow 64$) &
            conv3$\times$3 ($128 \rightarrow 128$) &
            conv3$\times$3 ($64 \rightarrow 64$) &
            conv3$\times$3 ($128 \rightarrow 128$) \\

            &&
            conv3$\times$3 ($64 \rightarrow 128$, ds) &
            conv3$\times$3 ($128 \rightarrow 256$, ds) &
            conv3$\times$3 ($64 \rightarrow 128$, ds) &
            conv3$\times$3 ($128 \rightarrow 256$, ds) \\

        \cline{2-6}
            & \multirow{2}{*}{$128 \times 128$} &
            conv3$\times$3 ($128 \rightarrow 128$) &
            conv3$\times$3 ($256 \rightarrow 256$) &
            conv3$\times$3 ($128 \rightarrow 128$) &
            conv3$\times$3 ($256 \rightarrow 256$) \\

            &&
            conv3$\times$3 ($128 \rightarrow 256$, ds) &
            conv3$\times$3 ($256 \rightarrow 512$, ds) &
            conv3$\times$3 ($128 \rightarrow 256$, ds) &
            conv3$\times$3 ($256 \rightarrow 512$, ds) \\

        \cline{2-6}
            & \multirow{2}{*}{$64 \times 64$} &
            conv3$\times$3 ($256 \rightarrow 256$) &
            conv3$\times$3 ($512 \rightarrow 512$) &
            conv3$\times$3 ($256 \rightarrow 256$) &
            conv3$\times$3 ($512 \rightarrow 512$) \\

            &&
            conv3$\times$3 ($256 \rightarrow 512$, ds) &
            conv3$\times$3 ($512 \rightarrow 512$, ds) &
            conv3$\times$3 ($256 \rightarrow 512$, ds) &
            conv3$\times$3 ($512 \rightarrow 512$, ds) \\

        \cline{2-6}
            & \multirow{2}{*}{$32 \times 32$} &
            conv3$\times$3 ($512 \rightarrow 512$) &
            conv3$\times$3 ($512 \rightarrow 512$) &
            conv3$\times$3 ($512 \rightarrow 512$) &
            conv3$\times$3 ($512 \rightarrow 512$) \\

            &&
            conv3$\times$3 ($512 \rightarrow 512$, ds) &
            conv3$\times$3 ($512 \rightarrow 512$, ds) &
            conv3$\times$3 ($512 \rightarrow 512$, ds) &
            conv3$\times$3 ($512 \rightarrow 512$, ds) \\

        \cline{2-6}
            & \multirow{2}{*}{$16 \times 16$} &
            conv3$\times$3 ($512 \rightarrow 512$) &
            conv3$\times$3 ($512 \rightarrow 512$) &
            conv3$\times$3 ($512 \rightarrow 512$) &
            conv3$\times$3 ($512 \rightarrow 512$) \\

            &&
            conv3$\times$3 ($512 \rightarrow 512$, ds) &
            conv3$\times$3 ($512 \rightarrow 512$, ds) &
            conv3$\times$3 ($512 \rightarrow 512$, ds) &
            conv3$\times$3 ($512 \rightarrow 512$, ds) \\

        \cline{2-6}
            & \multirow{2}{*}{$8 \times 8$} &
            conv3$\times$3 ($512 \rightarrow 512$) &
            conv3$\times$3 ($512 \rightarrow 512$) &
            conv3$\times$3 ($512 \rightarrow 512$) &
            conv3$\times$3 ($512 \rightarrow 512$) \\

            &&
            conv3$\times$3 ($512 \rightarrow 512$, ds) &
            conv3$\times$3 ($512 \rightarrow 512$, ds) &
            conv3$\times$3 ($512 \rightarrow 512$, ds) &
            conv3$\times$3 ($512 \rightarrow 512$, ds) \\

        \cline{2-6}
            & \multirow{3}{*}{$4 \times 4$} &
            conv3$\times$3 ($512 \rightarrow 512$) &
            conv3$\times$3 ($512 \rightarrow 512$) &
            conv3$\times$3 ($512 \rightarrow 512$) &
            conv3$\times$3 ($512 \rightarrow 512$) \\

            &&
            fc ($512\times4\times4 \rightarrow 1024$) &
            fc ($512\times4\times4 \rightarrow 1024$) &
            fc ($512\times4\times4 \rightarrow 1024$) &
            fc ($512\times4\times4 \rightarrow 1024$) \\

            &&
            dropout &
            dropout &
            dropout &
            dropout \\

        \hline
            \parbox[t]{2mm}{\multirow{24}{*}{\rotatebox[origin=c]{90}{synthesis}}} &
            \multirow{3}{*}{$4 \times 4$} &
            fc ($1024 \rightarrow 512\times4\times4$) &
            fc ($1024 \rightarrow 512\times4\times4$) &
            fc ($1024 \rightarrow 512\times4\times4$) &
            fc ($1024 \rightarrow 512\times4\times4$) \\

            && 
            conv3$\times$3 ($512 \rightarrow 512$) &
            conv3$\times$3 ($512 \rightarrow 512$) &
            conv3$\times$3 ($512 \rightarrow 512$) &
            conv3$\times$3 ($512 \rightarrow 512$) \\

            && 
            torgb ($512 \rightarrow 3$) &
            torgb ($512 \rightarrow 3$) &
            torgb ($512 \rightarrow 3$) &
            torgb ($512 \rightarrow 3$) \\

        \cline{2-6}
            & \multirow{3}{*}{$8 \times 8$} &
            conv3$\times$3 ($512 \rightarrow 512$, us) &
            conv3$\times$3 ($512 \rightarrow 512$, us) &
            conv3$\times$3 ($512 \rightarrow 512$, us) &
            conv3$\times$3 ($512 \rightarrow 512$, us) \\

            &&
            conv3$\times$3 ($512 \rightarrow 512$) &
            conv3$\times$3 ($512 \rightarrow 512$) &
            conv3$\times$3 ($512 \rightarrow 512$) &
            conv3$\times$3 ($512 \rightarrow 512$) \\

            && 
            torgb ($512 \rightarrow 3$) &
            torgb ($512 \rightarrow 3$) &
            torgb ($512 \rightarrow 3$) &
            torgb ($512 \rightarrow 3$) \\

        \cline{2-6}
            & \multirow{3}{*}{$16 \times 16$} &
            conv3$\times$3 ($512 \rightarrow 512$, us) &
            conv3$\times$3 ($512 \rightarrow 512$, us) &
            conv3$\times$3 ($512 \rightarrow 512$, us) &
            conv3$\times$3 ($512 \rightarrow 512$, us) \\

            &&
            conv3$\times$3 ($512 \rightarrow 512$) &
            conv3$\times$3 ($512 \rightarrow 512$) &
            conv3$\times$3 ($512 \rightarrow 512$) &
            conv3$\times$3 ($512 \rightarrow 512$) \\

            && 
            torgb ($512 \rightarrow 3$) &
            torgb ($512 \rightarrow 3$) &
            torgb ($512 \rightarrow 3$) &
            torgb ($512 \rightarrow 3$) \\

        \cline{2-6}
            & \multirow{3}{*}{$32 \times 32$} &
            conv3$\times$3 ($512 \rightarrow 512$, us) &
            conv3$\times$3 ($512 \rightarrow 512$, us) &
            conv3$\times$3 ($512 \rightarrow 512$, us) &
            conv3$\times$3 ($512 \rightarrow 512$, us) \\

            &&
            conv3$\times$3 ($512 \rightarrow 512$) &
            conv3$\times$3 ($512 \rightarrow 512$) &
            conv3$\times$3 ($512 \rightarrow 512$) &
            conv3$\times$3 ($512 \rightarrow 512$) \\

            && 
            torgb ($512 \rightarrow 3$) &
            torgb ($512 \rightarrow 3$) &
            torgb ($512 \rightarrow 3$) &
            torgb ($512 \rightarrow 3$) \\

        \cline{2-6}
            & \multirow{3}{*}{$64 \times 64$} &
            conv3$\times$3 ($512 \rightarrow 256$, us) &
            conv3$\times$3 ($512 \rightarrow 512$, us) &
            conv3$\times$3 ($512 \rightarrow 256$, us) &
            conv3$\times$3 ($512 \rightarrow 512$, us) \\

            &&
            conv3$\times$3 ($256 \rightarrow 256$) &
            conv3$\times$3 ($512 \rightarrow 512$) &
            conv3$\times$3 ($256 \rightarrow 256$) &
            conv3$\times$3 ($512 \rightarrow 512$) \\

            && 
            torgb ($256 \rightarrow 3$) &
            torgb ($512 \rightarrow 3$) &
            torgb ($256 \rightarrow 3$) &
            torgb ($512 \rightarrow 3$) \\

        \cline{2-6}
            & \multirow{3}{*}{$128 \times 128$} &
            conv3$\times$3 ($256 \rightarrow 128$, us) &
            conv3$\times$3 ($512 \rightarrow 256$, us) &
            conv3$\times$3 ($256 \rightarrow 128$, us) &
            conv3$\times$3 ($512 \rightarrow 256$, us) \\

            &&
            conv3$\times$3 ($128 \rightarrow 128$) &
            conv3$\times$3 ($256 \rightarrow 256$) &
            conv3$\times$3 ($128 \rightarrow 128$) &
            conv3$\times$3 ($256 \rightarrow 256$) \\

            && 
            torgb ($128 \rightarrow 3$) &
            torgb ($256 \rightarrow 3$) &
            torgb ($128 \rightarrow 3$) &
            torgb ($256 \rightarrow 3$) \\

        \cline{2-6}
            & \multirow{3}{*}{$256 \times 256$} &
            conv3$\times$3 ($128 \rightarrow 64$, us) &
            conv3$\times$3 ($256 \rightarrow 128$, us) &
            conv3$\times$3 ($128 \rightarrow 64$, us) &
            conv3$\times$3 ($256 \rightarrow 128$, us) \\

            &&
            conv3$\times$3 ($64 \rightarrow 64$) &
            conv3$\times$3 ($128 \rightarrow 128$) &
            conv3$\times$3 ($64 \rightarrow 64$) &
            conv3$\times$3 ($128 \rightarrow 128$) \\

            && 
            torgb ($64 \rightarrow 3$) &
            torgb ($128 \rightarrow 3$) &
            torgb ($64 \rightarrow 3$) &
            torgb ($128 \rightarrow 3$) \\

        \cline{2-6}
            & \multirow{3}{*}{$512 \times 512$} &
            &
            &
            conv3$\times$3 ($64 \rightarrow 32$, us) &
            conv3$\times$3 ($128 \rightarrow 64$, us) \\

            &&
            &
            &
            conv3$\times$3 ($32 \rightarrow 32$) &
            conv3$\times$3 ($64 \rightarrow 64$) \\

            && 
            &
            &
            torgb ($32 \rightarrow 3$) &
            torgb ($64 \rightarrow 3$) \\
        \hline
    \end{tabular}
}
\vspace{0.3cm}
\caption{
    The detail architecture of the encoder and synthesis network in our small and standard SH-GAN. All convolution layers are followed by the leaky ReLU activation with $\alpha=0.2$. Annotation \textit{ds} and \textit{us} means downsample and upsample. The $torgb$ layer is a conv1$\times$1 layer that converts image features into RGB images, which will be aggregated in parallel with the main architecture. For simplicity, we don't list out the mapping network, which is a sequential network with eight 512 to 512 fully connected layers plus ReLU. We don't list our Spectral Hint Unit (SHU) as well. SHU details can be found in the main article.
}
\label{table:architecture}
\end{table*}

\section{Masks}\label{appendix:masks}
\begin{figure*}[t]
\centering
    \hspace{0.05\textwidth}
    \begin{subfigure}[b]{0.28\textwidth}
    \centering
        \includegraphics[width=\textwidth]{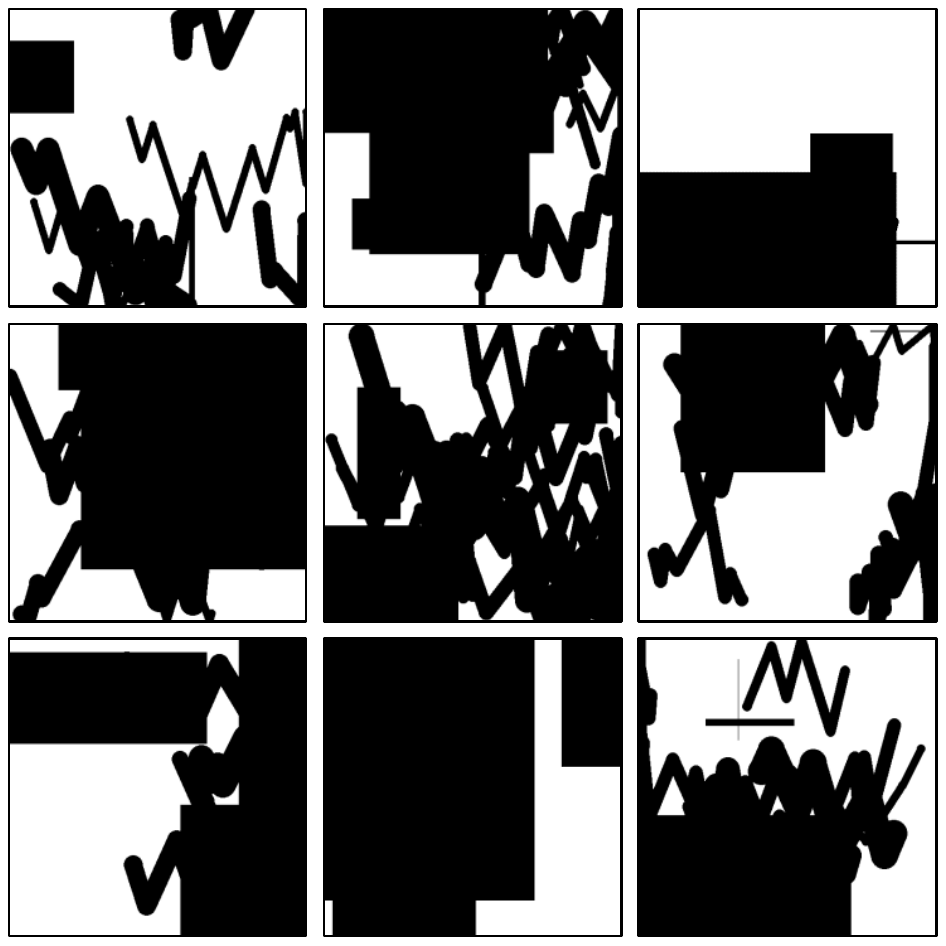}
    \subcaption{Free-form mask}
    \end{subfigure}
    \hfill
    \begin{subfigure}[b]{0.28\textwidth}
    \centering
        \includegraphics[width=\textwidth]{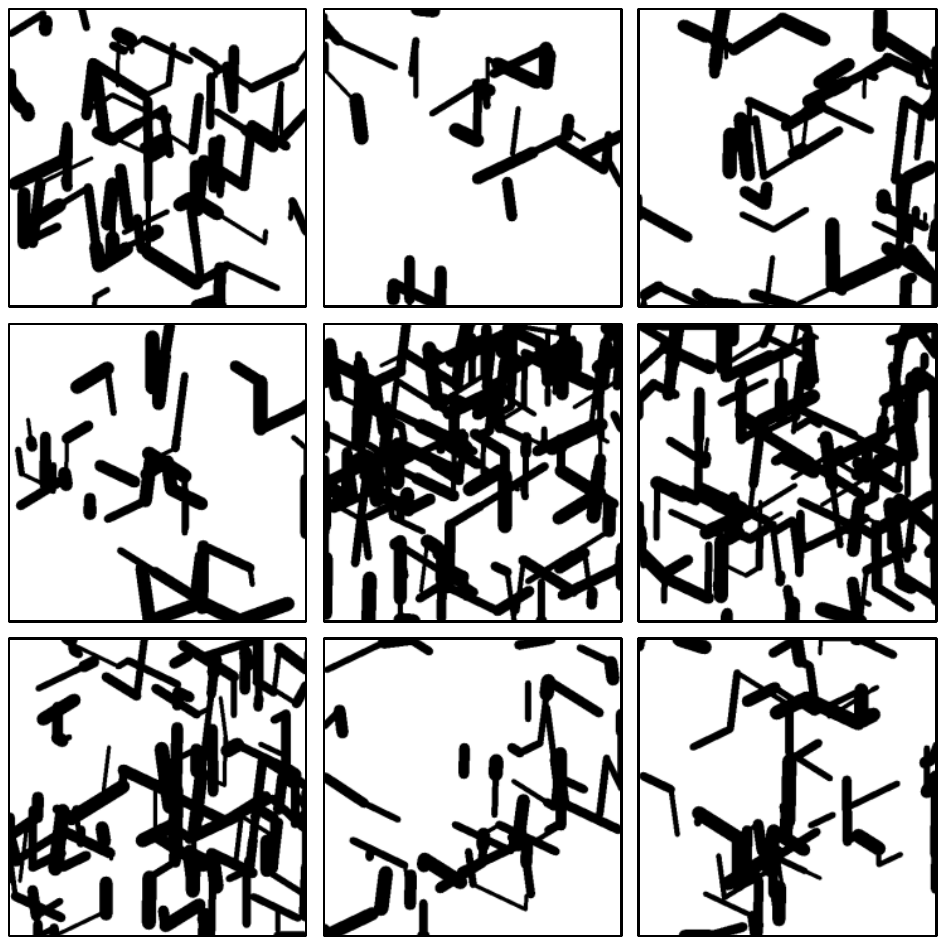}
    \subcaption{Narrow mask}
    \end{subfigure}
    \hfill
    \begin{subfigure}[b]{0.28\textwidth}
    \centering
        \includegraphics[width=\textwidth]{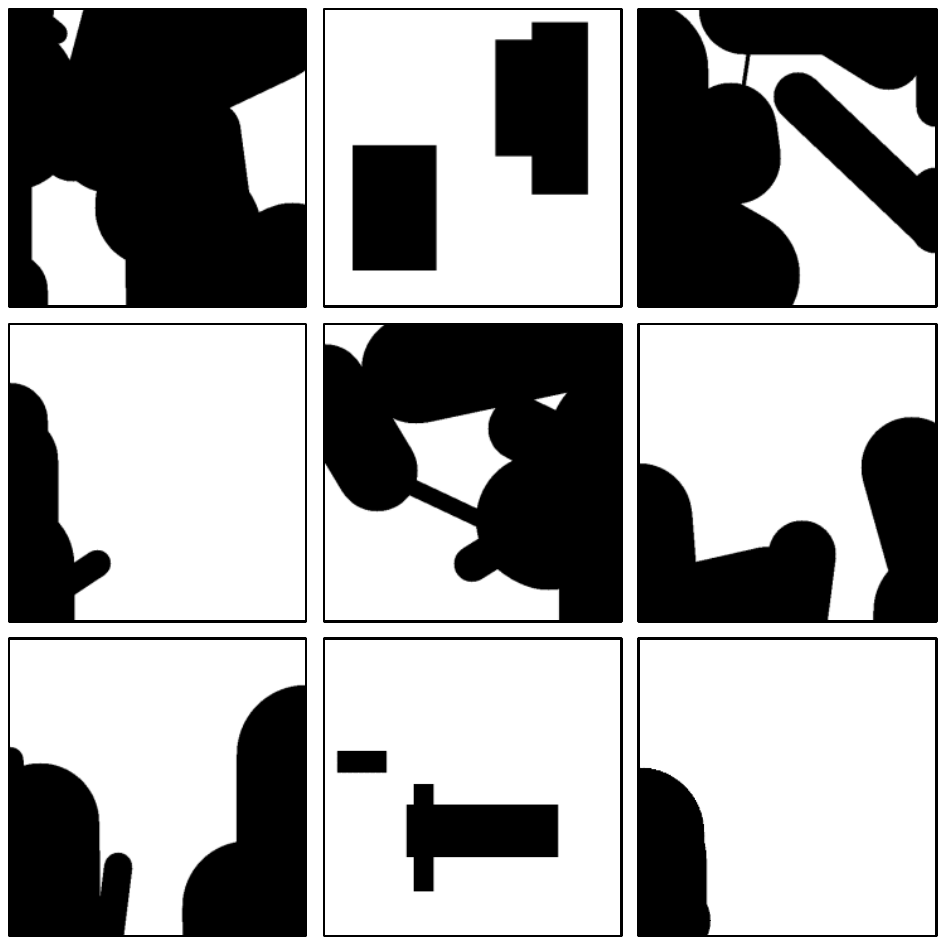}
    \subcaption{Wide mask}
    \end{subfigure}
    \hspace{0.05\textwidth}
\caption{The three types of masks we use in our experiments. The 1-value (white) represents known pixels and the 0-value (black) represents the unknown pixels.}
\vspace{-0.1cm}
\label{fig:masks}
\end{figure*}

As mentioned in section 3.4 of the main article, we followed the same algorithm as DeepFillv2~\cite{deepfillv2} and CoModGAN~\cite{comodgan} to generate free-form masks for training and evaluation. Besides, we adopted the LaMa-style~\cite{lama} narrow and wide masks to extend our evaluation beyond free-form masks scenarios. We use the official code downloaded from LaMa's Github to generate both narrow and wide masks. We list visualization for all three types of masks in figure~\ref{fig:masks}. As shown, narrow masks yield more and thinner strokes, while wide masks yield fewer and broader strokes. Both narrow and wide masks yield easier inpainting cases than the regular free-form masks we used in the main article. Moreover, we performed both quantitative and qualitative evaluations on these masks. Please see the following sections for more details.

\begin{table*}[t!]
\centering
\vspace{0.2cm}
\resizebox{\textwidth}{!}{
    \begin{tabular}{
            L{3.5cm}
            C{1.5cm}
            C{1.5cm}
            C{1.5cm}
            C{1.5cm}
            C{1.5cm}
            C{1.5cm}
            C{1.5cm}
            C{1.5cm}}
        \toprule
            & \multicolumn{4}{c}{\textbf{FFHQ 256}}
            & \multicolumn{4}{c}{\textbf{Places2 256}}
            \\
            & \multicolumn{2}{c}{\textbf{narrow}}
            & \multicolumn{2}{c}{\textbf{wide}}
            & \multicolumn{2}{c}{\textbf{narrow}}
            & \multicolumn{2}{c}{\textbf{wide}}
            \\
        \cmidrule(l{0.5em}r{0.5em}){2-3}
        \cmidrule(l{0.5em}r{0.5em}){4-5}
        \cmidrule(l{0.5em}r{0.5em}){6-7}
        \cmidrule(l{0.5em}r{0.5em}){8-9}
            \textbf{Method}
            & FID($\downarrow$)
            & LPIPS($\downarrow$) 
            & FID($\downarrow$)
            & LPIPS($\downarrow$) 
            & FID($\downarrow$)
            & LPIPS($\downarrow$)
            & FID($\downarrow$)
            & LPIPS($\downarrow$)
            \\
        \midrule
        CoModGan (small)
            & 2.0327
            & 0.0430
            & 2.5680
            & 0.1235
            & 1.7972
            & 0.0880
            & 3.4491
            & 0.2022
            \\
        CoModGan (official)
            & 1.7082
            & \best{0.0411}
            & 2.4859
            & 0.1224
            & 1.5214
            & 0.0857
            & 3.3955
            & 0.2016
            \\
        LaMa
            & 4.6266
            & 0.0418
            & 8.5166
            & \best{0.1169}
            & 11.4755
            & 0.0746
            & 12.6746
            & \best{0.1792}
            \\
        DeepFillV2
            & 8.5031
            & 0.0673
            & 14.7631
            & 0.1531
            & 15.0309
            & 0.0980
            & 20.6116
            & 0.2160
            \\
        CR-Fill
            & - 
            & - 
            & - 
            & -
            & 12.5929
            & 0.0935
            & 18.1343
            & 0.2036
            \\
        Onion-Conv
            & - 
            & - 
            & - 
            & -
            & 15.0021
            & 0.1107
            & 18.2985
            & 0.2239
            \\
        MADF
            & 1.5619
            & 0.0312
            & 7.3928
            & 0.1257
            & 10.4441
            & \best{0.0730}
            & 20.0603
            & 0.2023
            \\
        AOT-GAN
            & 2.1126
            & 0.0355
            & 15.6649
            & 0.1576
            & 9.9570
            & 0.0843
            & 26.5505
            & 0.2353
            \\
        \midrule
        (ours - small)
            & 2.0790
            & 0.0437
            & 2.5108
            & 0.1222
            & 1.5607
            & 0.0867
            & 3.0663
            & 0.2000
            \\
        (ours - regular)
            & \best{1.6847}
            & 0.0414
            & \best{2.3336}
            & 0.1214
            & \best{1.3084}
            & 0.0832
            & \best{2.7853}
            & 0.1982
            \\
        \bottomrule
    \end{tabular}
}
\vspace{0.2cm}
\caption{
    The performance on resolution 256 with LaMa-style~\cite{lama} narrow and wide masks.
}
\label{table:256_performance_lamastyle}
\end{table*}

\begin{table*}[t!]
\centering
\vspace{0.2cm}
\resizebox{\textwidth}{!}{
    \begin{tabular}{
            L{3.5cm}
            C{1.5cm}
            C{1.5cm}
            C{1.5cm}
            C{1.5cm}
            C{1.5cm}
            C{1.5cm}
            C{1.5cm}
            C{1.5cm}}
        \toprule
            & \multicolumn{4}{c}{\textbf{FFHQ 512}}
            & \multicolumn{4}{c}{\textbf{Places2 512}}
            \\
            & \multicolumn{2}{c}{\textbf{narrow}}
            & \multicolumn{2}{c}{\textbf{wide}}
            & \multicolumn{2}{c}{\textbf{narrow}}
            & \multicolumn{2}{c}{\textbf{wide}}
            \\
        \cmidrule(l{0.5em}r{0.5em}){2-3}
        \cmidrule(l{0.5em}r{0.5em}){4-5}
        \cmidrule(l{0.5em}r{0.5em}){6-7}
        \cmidrule(l{0.5em}r{0.5em}){8-9}
            \textbf{Method}
            & FID($\downarrow$)
            & LPIPS($\downarrow$) 
            & FID($\downarrow$)
            & LPIPS($\downarrow$)
            & FID($\downarrow$)
            & LPIPS($\downarrow$) 
            & FID($\downarrow$)
            & LPIPS($\downarrow$)
            \\
        \midrule
        CoModGan (small)
            & 1.4628
            & 0.0585
            & 2.8030
            & 0.1936
            & 1.2474
            & 0.0998
            & 4.8398
            & 0.2574
            \\
        CoModGan (official)
            & 1.2668
            & 0.0546
            & 2.7336
            & 0.1925
            & 1.0363
            & 0.0940
            & 4.5978
            & 0.2566
            \\
        LaMa
            & 2.3060
            & 0.0608
            & 11.6739
            & 0.2162
            & 1.2551
            & 0.0884
            & 7.6624
            & \best{0.2425}
            \\
        DeepFillV2
            & 9.0039
            & 0.1156
            & 19.9922
            & 0.2361
            & 2.0743
            & 0.1003
            & 18.3419
            & 0.2871
            \\
        CR-Fill
            & - 
            & - 
            & -
            & -
            & 1.7619
            & 0.0916
            & 14.8824
            & 0.2675
            \\
        Onion-Conv
            & - 
            & - 
            & -
            & -
            & 3.1910
            & 0.1149
            & 14.0048
            & 0.2999
            \\
        MADF
            & 1.4295
            & 0.0544
            & 11.2990
            & 0.2041
            & 1.0556
            & \best{0.0751}
            & 18.4953
            & 0.2532
            \\
        AOT-GAN
            & 1.7366
            & \best{0.0516}
            & 23.6286
            & 0.2458
            & 1.3204
            & 0.0808
            & 28.4871
            & 0.2950
            \\
        \midrule
        (ours - small)
            & 1.4070
            & 0.0592
            & 2.7658
            & 0.1934
            & 1.1583
            & 0.0993
            & 4.4051
            & 0.2570
            \\
        (ours - regular)
            & \best{1.2053}
            & 0.0550
            & \best{2.6009}
            & \best{0.1901}
            & \best{0.9307}
            & 0.0916
            & \best{3.9755}
            & 0.2532
            \\
        \bottomrule
    \end{tabular}
}
\vspace{0.2cm}
\caption{
    The performance on resolution 512 with LaMa-style~\cite{lama} narrow and wide masks.
}
\label{table:512_performance_lamastyle}
\end{table*}

\section{Extra Results}\label{appendix:extra_results}
As mentioned in the last section, we extensively tested the robustness of the good performance of SH-GAN beyond free-form mask scenarios, using LaMa-style~\cite{lama} narrow and wide masks. All quantitative results are listed in Table~\ref{table:256_performance_lamastyle} and~\ref{table:512_performance_lamastyle}. Like in the main article, we tested all approaches with resolutions 256 and 512, using dataset FFHQ~\cite{stylegan1} and Places2~\cite{places2}. As shown in the tables, our SH-GAN still outperforms other methods in terms of FID in all experiments, demonstrating that SH-GAN is effective under a wide variety of inpainting cases.

\clearpage
\section{Visualization}\label{appendix:visualization}
\begin{figure*}[t]
    \centering
    \includegraphics[width=0.85\textwidth]{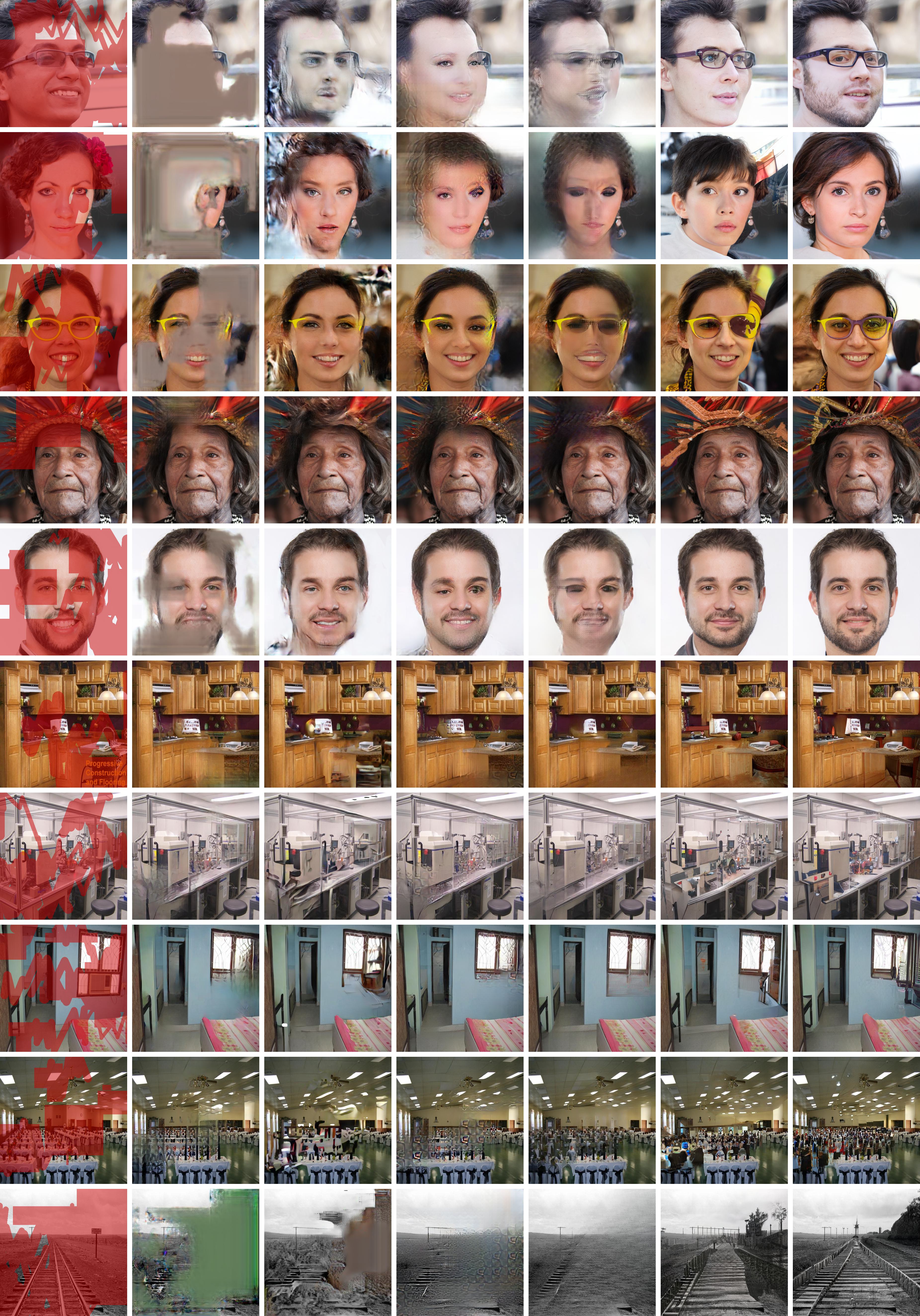}
    \includegraphics[width=0.85\textwidth]{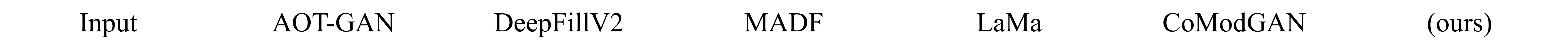}
    \caption{More qualitative examples between prior approaches and SH-GAN using free-form masks. Please zoom in for a better view.}
    \vspace{-0.3cm}
    \label{fig:qcompare_freeform}
\end{figure*}

\begin{figure*}[t]
    \centering
    \includegraphics[width=0.85\textwidth]{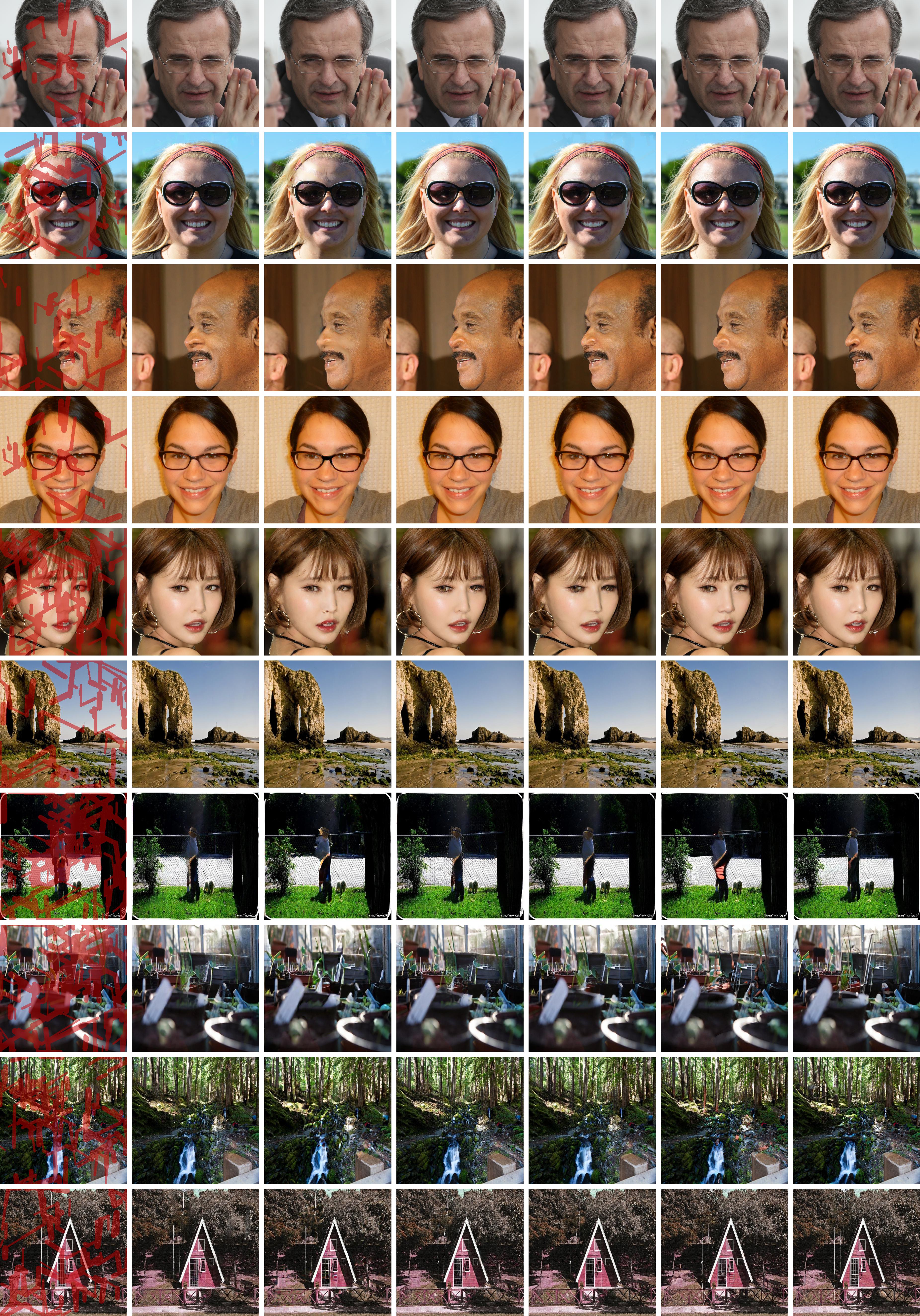}
    \includegraphics[width=0.85\textwidth]{figure-supp/qcompare_label.pdf}
    \caption{Qualitative examples between prior approaches and SH-GAN using narrow masks. Please zoom in for a better view.}
    \vspace{-0.3cm}
    \label{fig:qcompare_narrow}
\end{figure*}

\begin{figure*}[t]
    \centering
    \includegraphics[width=0.85\textwidth]{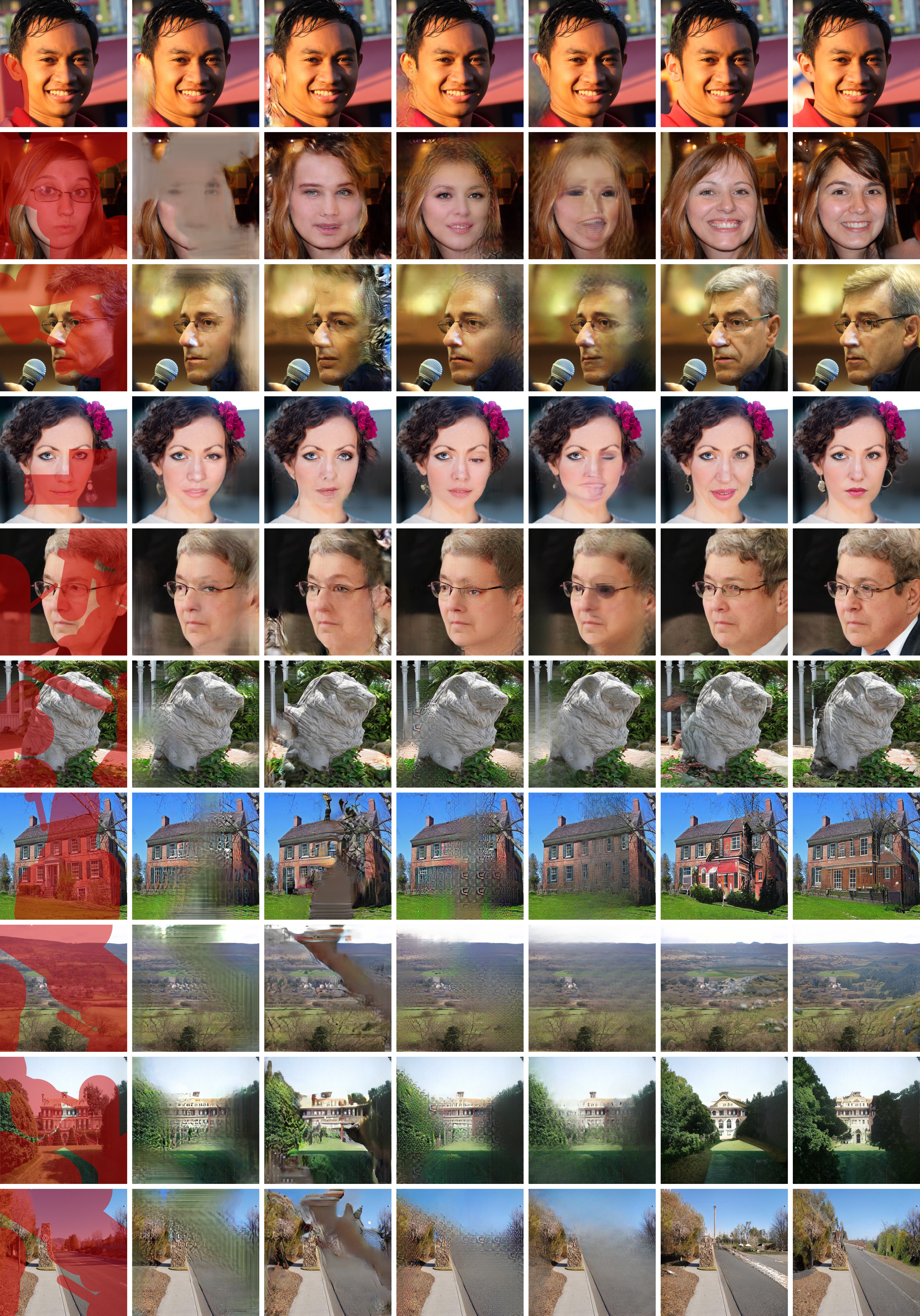}
    \includegraphics[width=0.85\textwidth]{figure-supp/qcompare_label.pdf}
    \caption{Qualitative examples between prior approaches and SH-GAN using wide masks. Please zoom in for a better view.}
    \vspace{-0.3cm}
    \label{fig:qcompare_wide}
\end{figure*}

In addition to the demo we showed in the main article, we generated more images by our SH-GAN and other approaches and qualitative compared them side by side in Figure~\ref{fig:qcompare_freeform}. Other than that, we also generate images using LaMa-style~\cite{lama} narrow and wide mask and listed them accordingly in Figure~\ref{fig:qcompare_narrow} and~\ref{fig:qcompare_wide}. 




\vspace{0.5cm}\label{appendix:controllable_editing}
\section{Controllable Editing}
\begin{figure}[h]
    \centering
    \includegraphics[width=0.8\columnwidth]{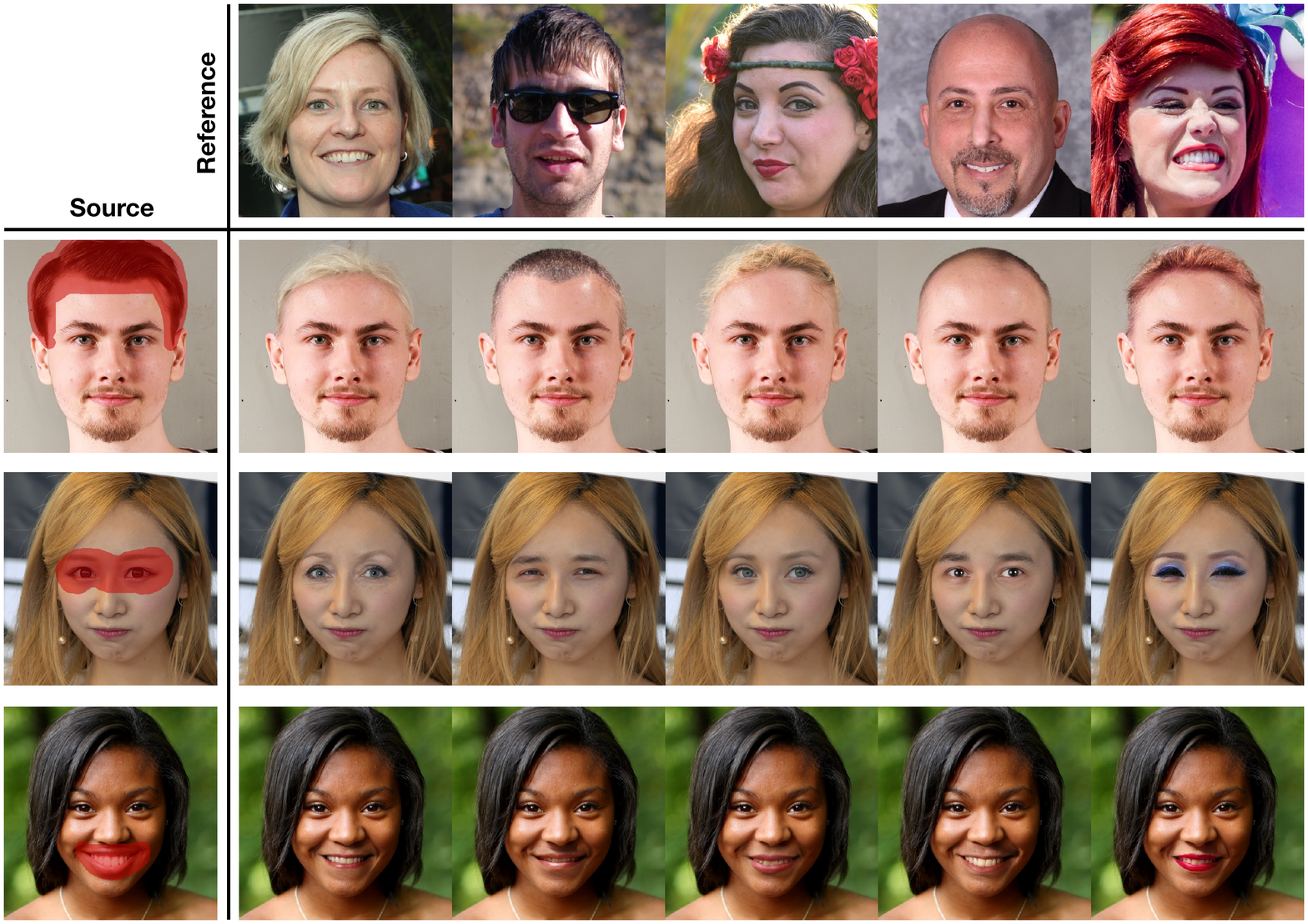}
    \caption{Controllable editing on different source and reference images from FFHQ.}
    \label{fig:editing}
\end{figure}

Controllable editing is one downstream application with excellent potential for SH-GAN in production. The definition of the task is to fill the missing region of an image $I$ with guidance from a reference image $I_R$. During inference, we pass both $I$, $I_R$, and the mask into the encoder. We then modulate our synthesis network using the global vector from $I_R$, and connect other intermediate features from $I$. Remind that this application is also feasible for CoModGAN~\cite{comodgan} but not for LaMa~\cite{lama} and other approaches. The results are shown in Figure~\ref{fig:editing}, in which we successfully transit parts of the reference face into the designated regions.

\end{appendices}

\end{document}